%% file: main.tex
\documentclass[10pt,twocolumn,letterpaper]{article}

\usepackage{cvpr}
\definecolor{cvprblue}{rgb}{0.21,0.49,0.74}
\usepackage[pagebackref,breaklinks,colorlinks,citecolor=cvprblue]{hyperref}
\usepackage{amsmath}
\usepackage{amssymb}
\usepackage{booktabs,caption}
\usepackage{multirow}
\usepackage{bm}
\usepackage{dsfont}
\usepackage[linesnumbered,ruled,vlined]{algorithm2e}
\usepackage{makecell}
\usepackage[flushleft]{threeparttable}
\usepackage{enumitem}
\usepackage[accsupp]{axessibility} 
\usepackage{siunitx}
\usepackage{graphicx}

\graphicspath{{figures/}}
\usepackage[capitalize]{cleveref}
\crefname{section}{Sec.}{Secs.}
\Crefname{section}{Section}{Sections}
\Crefname{table}{Table}{Tables}
\crefname{table}{Tab.}{Tabs.}

\def\confName{CVPR}
\def\confYear{2025}

\begin{document}

\title{NTIRE 2025 Challenge on Low Light Image Enhancement: Methods and Results}
\author{ \small Xiaoning Liu$^*$ \and \small
Zongwei Wu$^*$ \and \small
Florin-Alexandru Vasluianu$^*$ \and \small
Hailong Yan$^*$ \and \small
Bin Ren$^*$ \and \small
Yulun Zhang$^*$ \and \small
Shuhang Gu$^*$ \and \small
Le Zhang$^*$ \and \small
Ce Zhu$^*$ \and \small
Radu Timofte$^*$\and \small
Kangbiao Shi \and \small
Yixu Feng \and \small
Tao Hu \and \small
Yu Cao \and \small
Peng Wu \and \small
Yijin Liang \and \small
Yanning Zhang \and \small
Qingsen Yan \and \small
Han Zhou \and  \small
Wei Dong \and  \small
Yan Min \and  \small
Mohab Kishawy \and  \small
Jun Chen \and  \small
Pengpeng Yu  \and  \small
Anjin Park \and  \small
Seung-Soo Lee  \and  \small
Young-Joon Park  \and  \small
Zixiao Hu \and \small
Junyv Liu \and  \small
Huilin Zhang \and  \small
Jun Zhang \and  \small
Fei Wan \and  \small
Bingxin Xu \and  \small
Hongzhe Liu \and  \small
Cheng Xu \and  \small
Weiguo Pan \and   \small
Songyin Dai \and  \small
Xunpeng Yi \and  \small
Qinglong Yan \and  \small
Yibing Zhang \and \small
Jiayi Ma \and  \small
Changhui Hu \and \small
Kerui Hu \and \small
Donghang Jing \and \small
Tiesheng Chen \and \small
Zhi Jin \and \small
Hongjun Wu \and \small
Biao Huang \and \small
Haitao Ling \and \small
Jiahao Wu \and \small
Dandan Zhan \and \small
G Gyaneshwar Rao \and \small
Vijayalaxmi Ashok Aralikatti \and \small
Nikhil Akalwadi \and \small
Ramesh Ashok Tabib \and \small
Uma Mudenagudi \and \small
Ruirui Lin \and \small
Guoxi Huang\and \small
Nantheera Anantrasirichai \and \small
Qirui Yang \and \small
Alexandru Brateanu \and \small
Ciprian Orhei \and \small
Cosmin Ancuti \and \small
Daniel Feijoo \and \small
Juan C. Benito \and \small
Álvaro García \and \small
Marcos V. Conde \and \small
Yang Qin \and \small
Raul Balmez \and \small
Anas M. Ali \and \small
Bilel Benjdira \and \small
Wadii Boulila \and \small
Tianyi Mao \and \small
Huan Zheng \and \small
Yanyan Wei \and \small
Shengeng Tang \and \small
Dan Guo \and \small
Zhao Zhang \and \small
Sabari Nathan \and \small
K Uma \and \small
A Sasithradevi \and \small
B Sathya Bama \and \small
S. Mohamed Mansoor Roomi \and \small
Ao Li \and \small
Xiangtao Zhang \and \small
Zhe Liu \and \small
Yijie Tang \and \small
Jialong Tang \and \small
Zhicheng Fu \and \small
Gong Chen \and \small
Joe Nasti \and \small
John Nicholson \and \small
Zeyu Xiao \and \small
Zhuoyuan Li \and \small
Ashutosh Kulkarni \and \small
Prashant W. Patil \and \small
Santosh Kumar Vipparthi \and \small
Subrahmanyam Murala \and \small
Duan Liu \and \small
Weile Li \and \small
Hangyuan Lu \and \small
Rixian Liu \and \small
Tengfeng Wang \and \small
Jinxing Liang \and \small
Chenxin Yu
}
\maketitle
\let\thefootnote\relax\footnotetext{$^*$ X. Liu, Z. Wu, H. Yan, F. Vasluianu, B. Ren, Y. Zhang, S. Gu, L. Zhang, C. Zhu and R. Timofte were the challenge organizers, while the other authors participated in the challenge. Each team described its own method in the report, shortened by the organizers to meet 8 page criteria. Appendix~\ref{sec:teams} contains the teams, affiliations and architectures if available. NTIRE 2025 webpage:~\url{https://cvlai.net/ntire/2025}. Code:~\url{https://github.com/AVC2-UESTC/NTIRE2025-LLIE}.} 
\begin{abstract}
This paper presents a comprehensive review of the NTIRE 2025 Low-Light Image Enhancement (LLIE) Challenge, highlighting the proposed solutions and final outcomes. The objective of the challenge is to identify effective networks capable of producing brighter, clearer, and visually compelling images under diverse and challenging conditions. A remarkable total of 762 participants registered for the competition, with 28 teams ultimately submitting valid entries. This paper thoroughly evaluates the state-of-the-art advancements in LLIE, showcasing the significant progress.
\end{abstract}

\section{Introduction}
\label{sec:introduction}

Low-Light Image Enhancement (LLIE) aims to improve visibility and contrast across a wide range of low-light conditions.~In addition to enhancing brightness, LLIE seeks to address issues such as noise, artifacts, and color distortion, which are prevalent in dark scenes or arise during the illumination correction process.

Building upon the success of the NTIRE 2024 LLIE Challenge \cite{liu2024ntire}, we launched a new iteration at the NTIRE 2025 workshop. The 2025 challenge continues to encourage innovation by proposing solutions that significantly enhance image quality under complex low-light scenarios.

The goals of the challenge are threefold: (1) to drive research progress in the field of LLIE, (2) to enable systematic comparison of emerging methodologies, and (3) to provide a platform for academic and industrial participants to exchange ideas and explore potential collaborations.

Following a similar setup to the NTIRE 2024 edition \cite{liu2024ntire}, the dataset comprises a diverse set of scenarios under varying lighting conditions, including dim scenes, severe darkness, backlighting, non-uniform illumination, and both indoor and outdoor night scenes, with image resolutions reaching 4K and beyond.~The dataset includes 219 training scenes, along with 46 for validation and 30 for testing. Ground-truth (GT) images for both the validation and testing sets were kept hidden from participants.~Detailed dataset specifications will be published in future work.


This challenge is one of the NTIRE 2025 Workshop associated challenges on:~ambient lighting normalization~\cite{ntire2025ambient}, reflection removal in the wild~\cite{ntire2025reflection}, shadow removal~\cite{ntire2025shadow}, event-based image deblurring~\cite{ntire2025event}, image denoising~\cite{ntire2025denoising}, XGC quality assessment~\cite{ntire2025xgc}, UGC video enhancement~\cite{ntire2025ugc}, night photography rendering~\cite{ntire2025night}, image super-resolution (x4)~\cite{ntire2025srx4}, real-world face restoration~\cite{ntire2025face}, efficient super-resolution~\cite{ntire2025esr}, HR depth estimation~\cite{ntire2025hrdepth}, efficient burst HDR and restoration~\cite{ntire2025ebhdr}, cross-domain few-shot object detection~\cite{ntire2025cross}, short-form UGC video quality assessment and enhancement~\cite{ntire2025shortugc,ntire2025shortugc_data}, text to image generation model quality assessment~\cite{ntire2025text}, day and night raindrop removal for dual-focused images~\cite{ntire2025day}, video quality enhancement for video conferencing~\cite{ntire2025vqe}, low light image enhancement, light field super-resolution~\cite{ntire2025lightfield}, restore any image model (RAIM) in the wild~\cite{ntire2025raim}, raw restoration and super-resolution~\cite{ntire2025raw} and raw reconstruction from RGB on smartphones~\cite{ntire2025rawrgb}.

\section{Tracks and Competition}

\noindent\textbf{Ranking criteria.}~To evaluate the submissions, we use conventional metrics such as PSNR, SSIM, LPIPS, and NIQE. As shown in~\cref{tbl:ntire24_results}, the ``Final Rank'' is a composite metric derived from a weighted combination of PSNR (50\%), SSIM (50\%), LPIPS (40\%), and NIQE (20\%).

\noindent\textbf{Challenge phases.~}
\textit{(1) Development and validation phase:} Participants were provided with 219 training image pairs and 46 validation inputs from our custom dataset. The ground-truth images for the validation set were not shared. Participants could submit their enhanced results to an evaluation server, which computed PSNR and SSIM scores and provided real-time feedback.~\textit{(2) Testing phase:} Participants received 30 low-light test images, again without access to the corresponding ground-truth images.~Submissions, including enhanced results, accompanying code, and a factsheet, were uploaded to the Codalab evaluation server and shared with the organizers. The organizers verified and the final results. Top-performing teams were required to submit training scripts to ensure reproducibility.

\input{final_results}

\section{Challenge Methods and Teams}
\label{sec:methods_and_teams}

The results of the low light enhancement challenge are detailed in~\cref{tbl:ntire24_results}, which evaluates and ranks the performances of 28 teams.~One team (JHC-Info) fails to provide the checkpoint for reproducibility within the competition period, hence excluded ranking. Some others team provide all the fact sheet but didnt particpate in the challenge report. These works are highlight by \textcolor{gray}{Gray}

\input{team01_NWPU_HVI/main}

\input{team02_Imagine/main}
\input{team03_pengpeng_yu/main}
\input{team04_DAVIS_K/main}
\input{team05_SoloMan/main}
\input{team06_Smartdsp/main}
\input{team07_Smart210/main}

\input{team08_WHU_MVP/main}
\input{team09_BUPTMM/main}
\input{team10_NJUPTIPR/main}
\input{team11_SYSU_FVL_T2/main}
\input{team12_KLETech_CEVI/main}
\input{team13_Ensemble_KNights/main}
\input{team14_MRT_LLIE/main}
\input{team15_SynLLIE/main}
\input{team16_CidautAI/main}
\input{team17_Huabujianye/main}
\input{team18_no_way_no_lay/main}
\input{team19_Lux_Themps/main}
\input{team20_PSU_TEAM/main}

\input{team21_hfut_lvgroup/main}
\input{team22_ImageLab/main}
\input{team23_AVC2/main}

\input{team24_LR_LL/main}
\input{team25_X_L/main}
\input{team26_Team_IITRPR/main}
\input{team27_CV_SVNIT/main}
\input{team28_JHC_INFO/main}

\section*{Acknowledgements}
This work was partially supported by the Humboldt Foundation. We thank the NTIRE 2025 sponsors: ByteDance, Meituan, Kuaishou, and University of Wurzburg (Computer Vision Lab).

{\small
\bibliographystyle{ieee_fullname}
\bibliography{main}
}

\clearpage
\appendix

\section{Teams and Affiliations}
\label{sec:teams}

\subsection*{NTIRE 2025 team}
\noindent\textit{\textbf{Title: }} NTIRE 2025 Low Light Image Enhancement Challenge\\
\noindent\textit{\textbf{Members: }} \\
Xiaoning Liu$^1$ (\href{mailto:liuxiaoning2016@sina.com}{liuxiaoning2016@sina.com}),\\
Zongwei Wu$^2$
(\href{mailto:zongwei.wu@uni-wuerzburg.de}{zongwei.wu@uni-wuerzburg.de}),\\
Florin Vasluianu$^2$
(\href{mailto:florin-alexandru.vasluianu@uni-wuerzburg.de}{florin-alexandru.vasluianu@uni-wuerzburg.de}),\\
Hailong Yan$^1$
(\href{mailto:yhl00825@163.com}{yhl00825@163.com}),\\
Bin Ren$^4$
(\href{mailto:bin.ren@unitn.it}{bin.ren@unitn.it}),\\
Yulun Zhang$^3$ (\href{mailto:yulzhang@ethz.ch}{yulun100@gmail.com}),\\
Shuhang Gu$^1$ (\href{mailto:shuhanggu@gmail.com}{shuhanggu@gmail.com}),\\
Le Zhang$^1$
(\href{mailto:lezhang@uestc.edu.cn}{lezhang@uestc.edu.cn}),\\
Ce Zhu$^1$
(\href{mailto:eczhu@uestc.edu.cn}{eczhu@uestc.edu.cn}),\\
Radu Timofte$^{2}$ (\href{mailto:radu.timofte@uni-wuerzburg.de}{radu.timofte@uni-wuerzburg.de})\\
\noindent\textit{\textbf{Affiliations:}}\\
$^1$~University of Electronic Science and Technology of China, China\\
$^2$~Computer Vision Lab, University of W\"urzburg, Germany\\
$^3$~Shanghai Jiao Tong University, China\\
$^4$~University of Trento, Italy\\
\vspace{-2mm}
\input{team01_NWPU_HVI/affiliation}
\vspace{-2mm}
\input{team02_Imagine/affiliation}
\vspace{-2mm}
\input{team03_pengpeng_yu/affiliation}
\vspace{-2mm}
\input{team04_DAVIS_K/affiliation}
\vspace{-2mm}
\input{team05_SoloMan/affiliation}

\vspace{-2mm}
\input{team06_Smartdsp/affiliation}
\vspace{-2mm}
\input{team07_Smart210/affiliation}
\vspace{-2mm}
\input{team08_WHU_MVP/affiliation}

\vspace{-2mm}
\input{team09_BUPTMM/affiliation}
\vspace{-2mm}
\input{team10_NJUPTIPR/affiliation}
\vspace{-2mm}
\input{team11_SYSU_FVL_T2/affiliation}
\vspace{-2mm}
\input{team12_KLETech_CEVI/affiliation}
\vspace{-2mm}
\input{team13_Ensemble_KNights/affiliation}
\vspace{-2mm}
\input{team14_MRT_LLIE/affiliation}
\vspace{-2mm}
\input{team15_SynLLIE/affiliation}
\vspace{-2mm}
\input{team16_CidautAI/affiliation}
\vspace{-2mm}
\input{team17_Huabujianye/affiliation}
\vspace{-2mm}
\input{team18_no_way_no_lay/affiliation}
\vspace{-2mm}
\input{team19_Lux_Themps/affiliation}
\vspace{-2mm}
\input{team20_PSU_TEAM/affiliation}
\vspace{-2mm}
\input{team21_hfut_lvgroup/affiliation}
\vspace{-2mm}
\input{team22_ImageLab/affiliation}

\vspace{-2mm}
\input{team23_AVC2/affiliation}
\vspace{-2mm}
\input{team24_LR_LL/affiliation}
\vspace{-2mm}
\input{team25_X_L/affiliation}
\vspace{-2mm}
\input{team26_Team_IITRPR/affiliation}
\vspace{-2mm}
\input{team27_CV_SVNIT/affiliation}
\vspace{-2mm}
\input{team28_JHC_INFO/affiliation}
\newpage
\begin{table}
\centering
\caption{Training configurations for different models. LR represents learning rate in training process.}
\vspace{-3mm}
\label{tab:training_config}
\resizebox{\linewidth}{!}{
\begin{tabular}{c|c|c}
\toprule
\textbf{Model} & \textbf{Hardware}  & \textbf{Initial/Final LR} \\
\midrule
RetinexFormer \cite{retinexformer} & Tesla A100  & $2\times10^{-4} \rightarrow 1\times10^{-6}$ \\
ESDNet \cite{yu2022towards} & Tesla A100  & $2\times10^{-4} \rightarrow 1\times10^{-6}$  \\
CIDNet \cite{yan2025hvi}& RTX 4090  & $1\times10^{-4} \rightarrow 1\times10^{-7}$ \\
\bottomrule
\end{tabular}}
\vspace{-5mm}
\end{table}
\vspace{-4mm}
\begin{figure}
	\centering
	\includegraphics[width=1\linewidth]{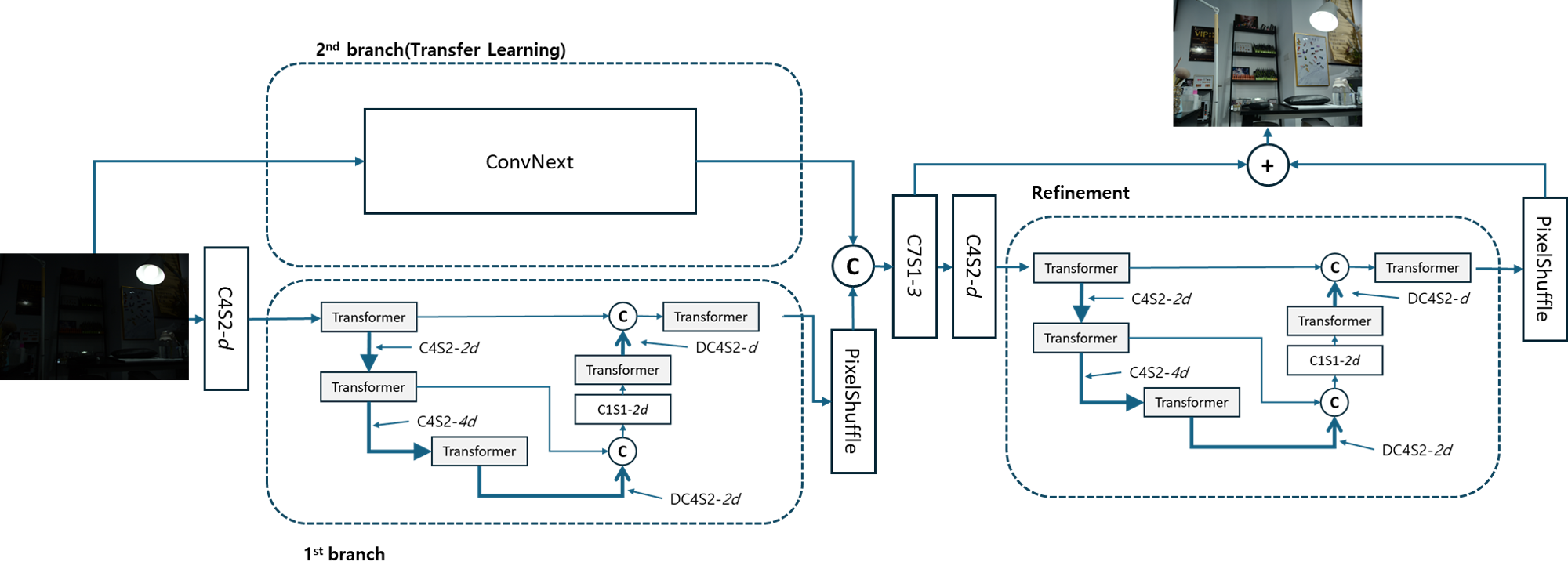}
	\vspace{-7mm}
	\caption{Architecture of Team DAVIS-K.}
	\vspace{-7mm}
	\label{fig:fig4-1}
\end{figure}
\begin{figure}
    \includegraphics[width=1\linewidth]{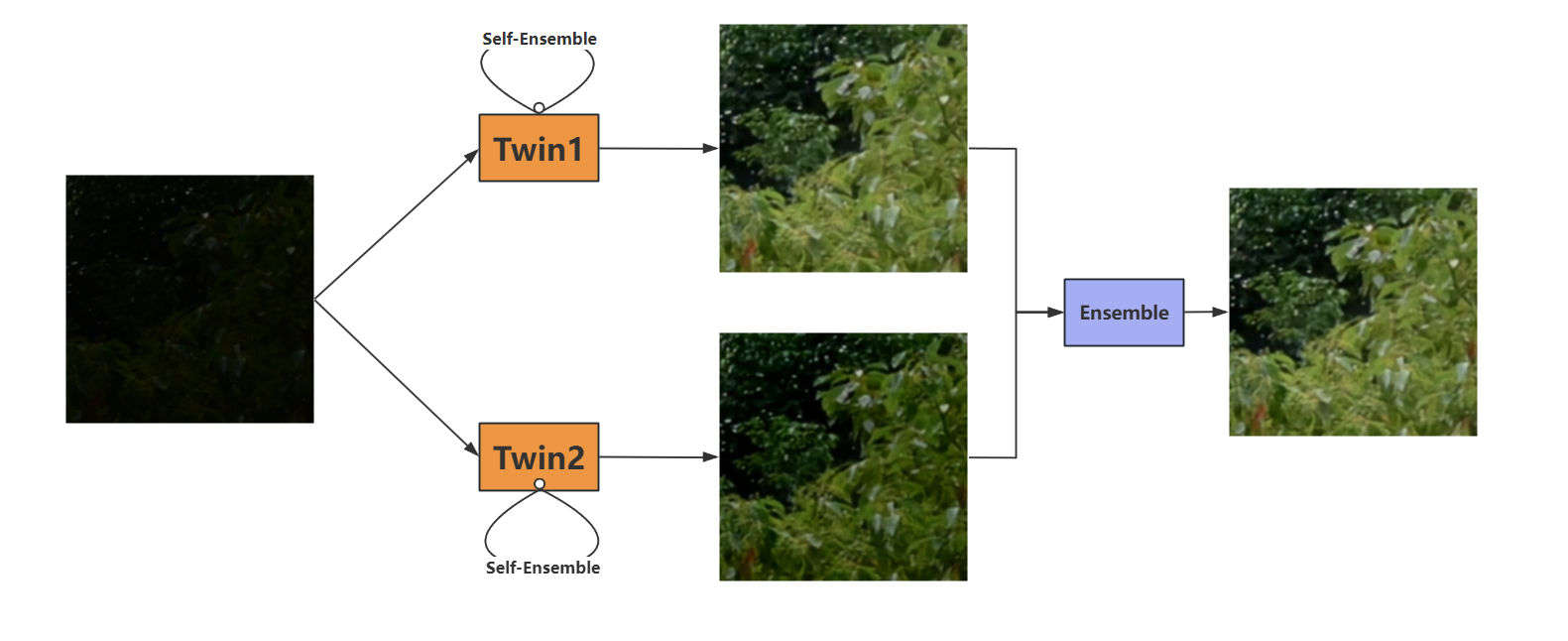}
    \caption{The pipeline of the ESDNet-Twins.}
    \vspace{-2mm}
    \label{fig:enter-label_pipnline}
\end{figure}
\begin{figure}
  \centering
 \includegraphics[width=\linewidth]{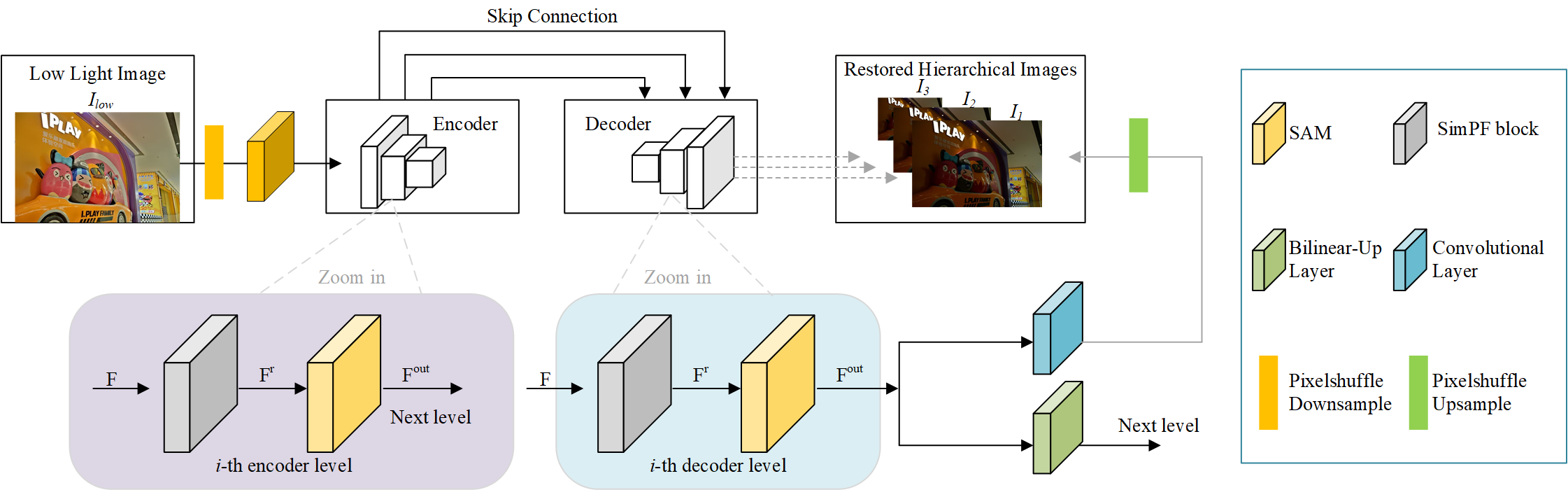}
\caption{ESDNet \cite{yu2022towards} with SimPFblock \cite{wan2024psc} for low light image enhancement.}
\label{fig7-1}%
\end{figure}
\begin{figure}
  \centering
 \includegraphics[width=\linewidth]{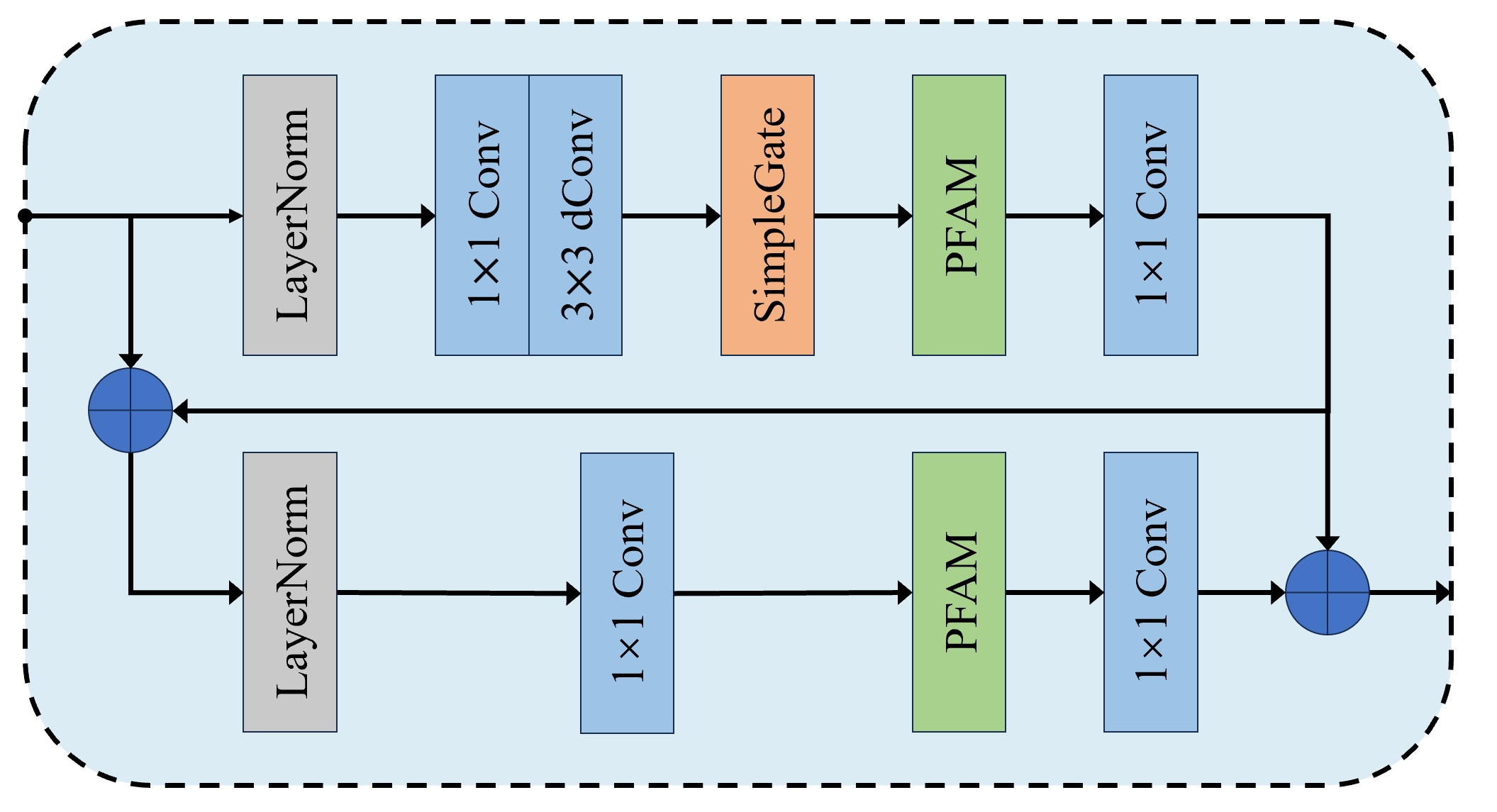}
\caption{Illustration of the SimPFblock \cite{wan2024psc}.}
\label{fig7-2}%
\end{figure}
\begin{figure}
  \centering
 \includegraphics[width=\linewidth]{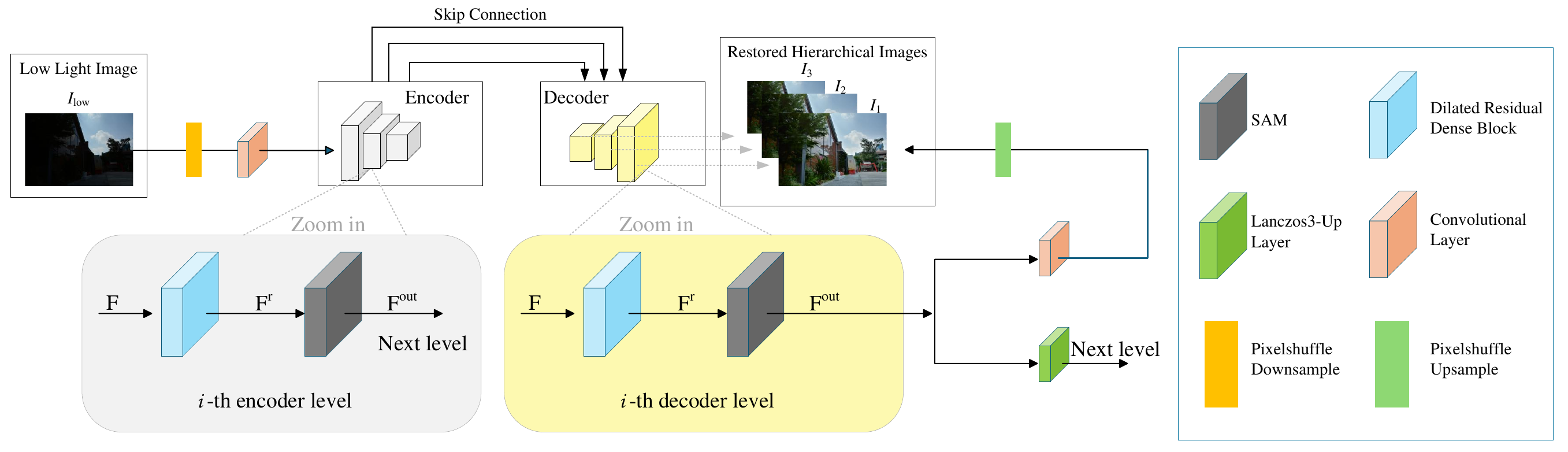}
\caption{The backbone network used in our method. (Reproduced from ESDNet \cite{yu2022towards}).}
\label{fig_esdnet}%
\end{figure}
\begin{figure}[!t]
    \centering
    \includegraphics[width=1\linewidth]{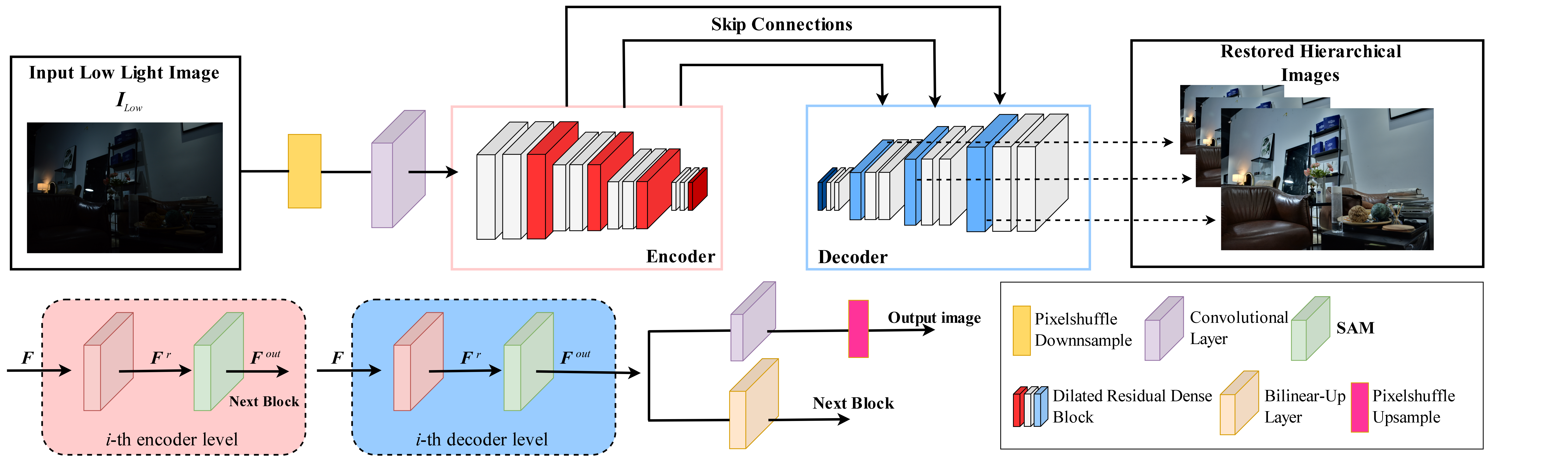}
    \caption{Architecture of the proposed ESDNet+.}
    \label{fig:esdnet+}
\end{figure}
\begin{figure}
    \centering
    \includegraphics[width=1\linewidth]{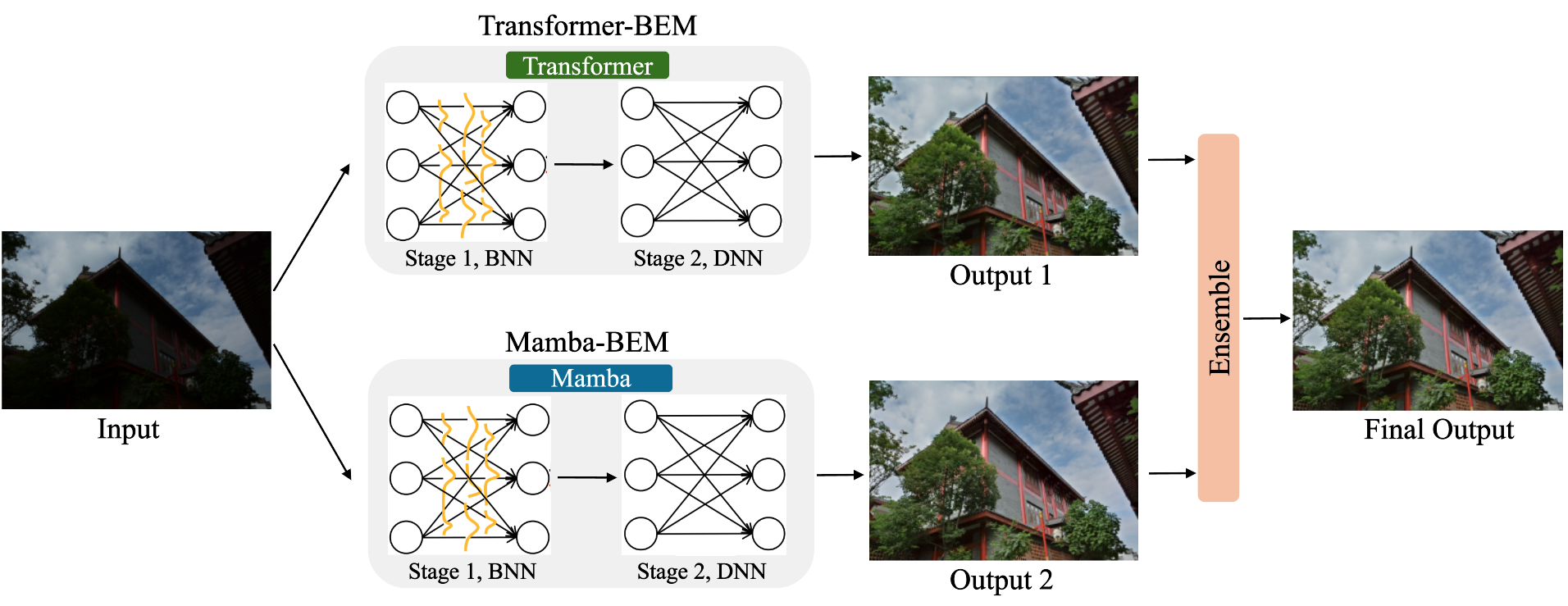}
    \caption{Overall framework of ENsemble BEM~\cite{huang2025bayesian} (EN-BEM).}
    \label{fig:enbem}
\end{figure}
\begin{figure}
    \centering
    \includegraphics[width=1.0\linewidth]{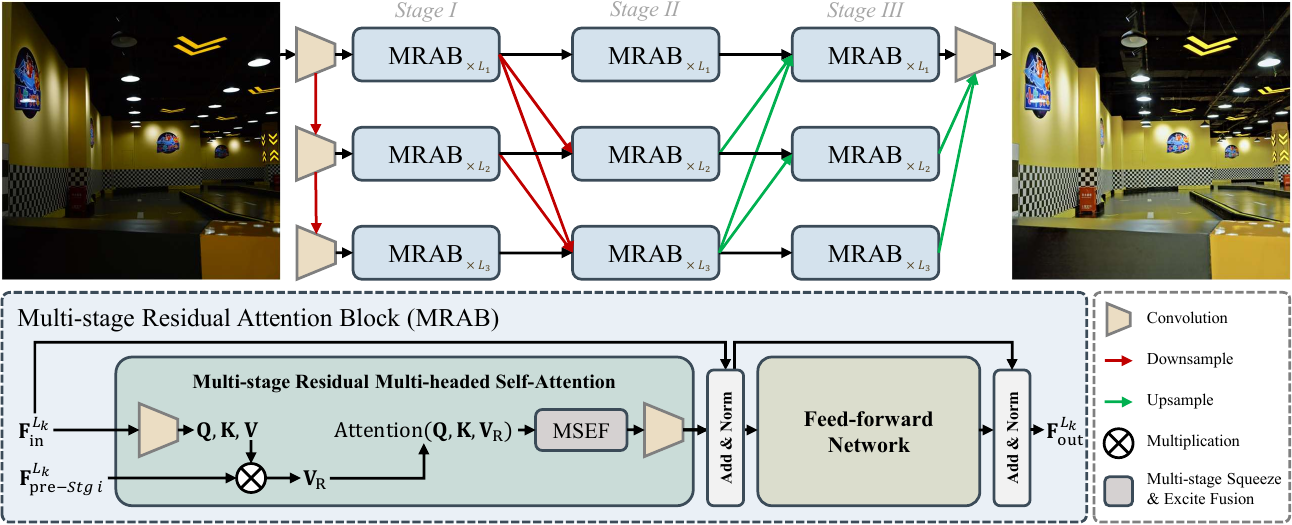}
    \vspace{-5pt}
    \caption{Overall framework of our Multi-stage Residual Transformer (\textbf{MRT}).}
    \vspace{-5pt}
    \label{fig:mrt_framework}
    \vspace{-5pt}
\end{figure}
\begin{figure}
\centering
\includegraphics[width=23em]{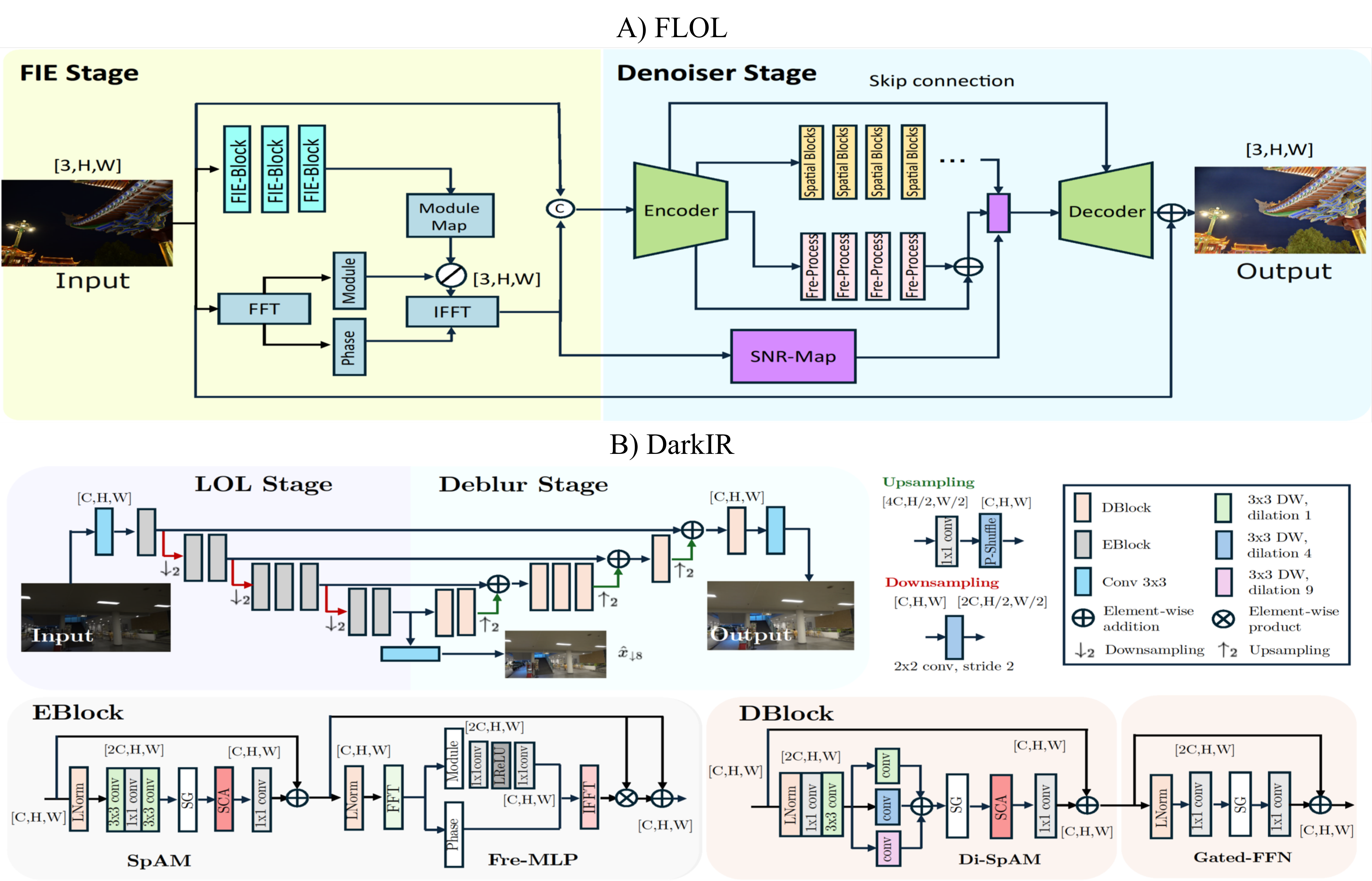}
\caption{Overview of the two original solutions. We implement Fourier frequency information in both cases. In A), we employ that feature to obtain a lightweight model capable of processing challenge images with a mean time of only \textbf{0.15 s per image} and obtaining \textbf{24.15 dB} and \textbf{0.82} of PSNR and SSIM, respectively. In B), we expose a more complex model which reaches better values in evaluation metrics such as PSNR, SSIM and LPIPS -- shown in Tab. \ref{tbl:ntire24_results}.}
\label{fig:method_cidautai}
\end{figure}
\begin{figure}
    \centering
    \includegraphics[width=0.95\linewidth]{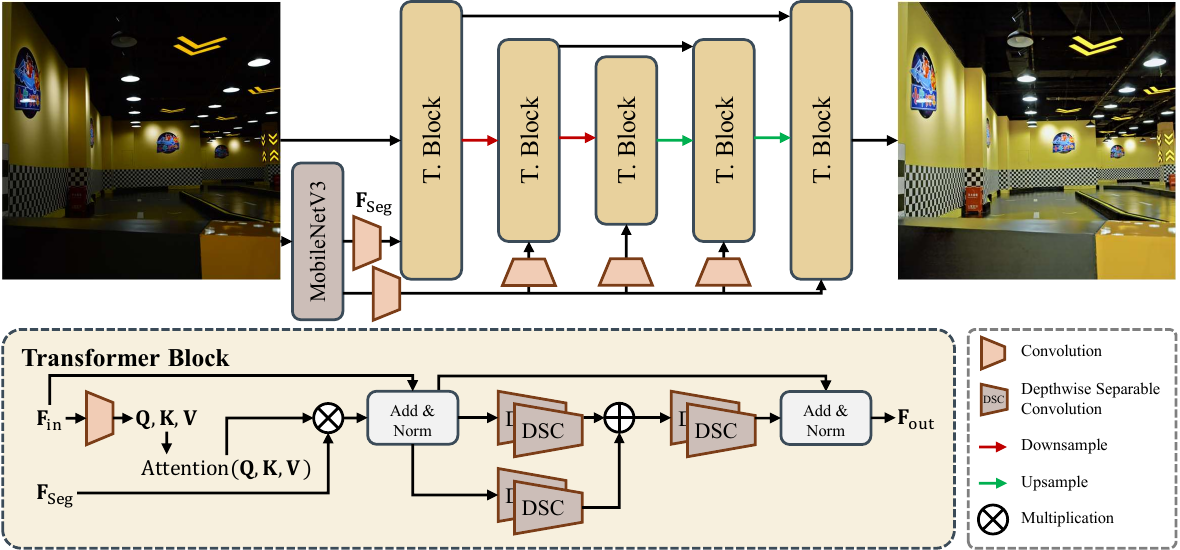}
    \caption{Overall framework of our Semantic-Aware Depthwise Vision Transformer (\textbf{SADe ViT}).}
    \label{fig:sade_framework}
    \vspace{-4mm}
\end{figure}
\begin{figure}
\centering
\includegraphics[width=23em]{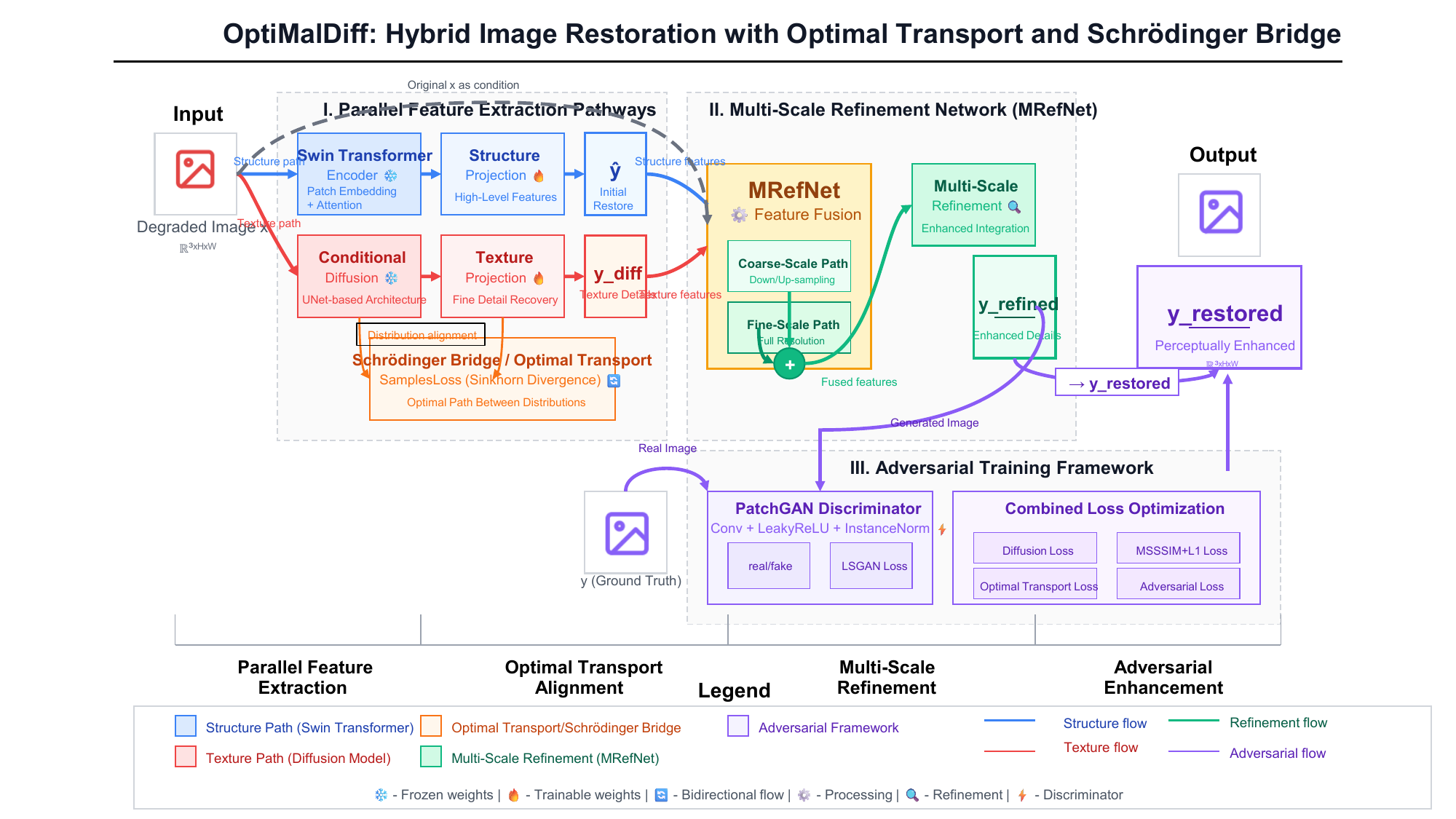}
\caption{Overview of the OptiMalDiff architecture combining Schrödinger Bridge diffusion, transformer-based feature extraction, and adversarial refinement.}
\label{fig:method_psu}
\vspace{-2mm}
\end{figure}
\begin{figure}
    \centering
    \includegraphics[width=\linewidth]{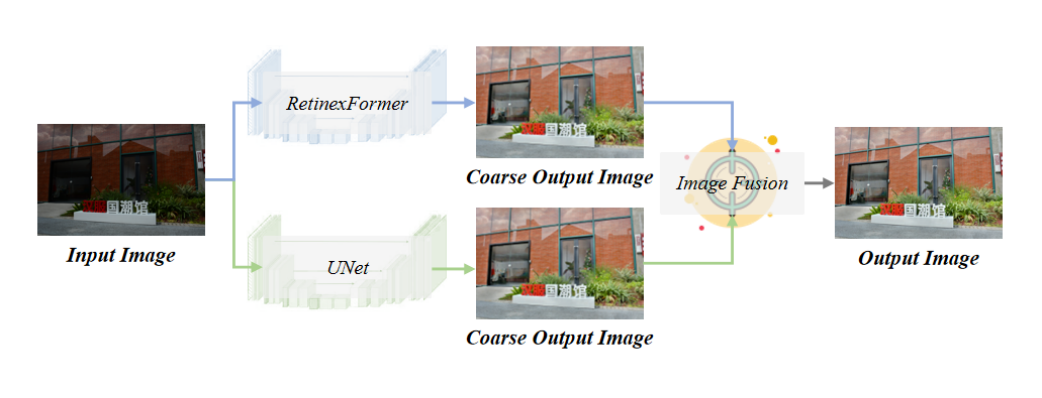}
    \caption{Overview of the proposed model fusion strategy for low-light image enhancement.}
    \label{fig:model_framework21}
    \vspace{-4mm}
\end{figure}
\begin{figure}
    \centering
    \includegraphics[width=\linewidth]{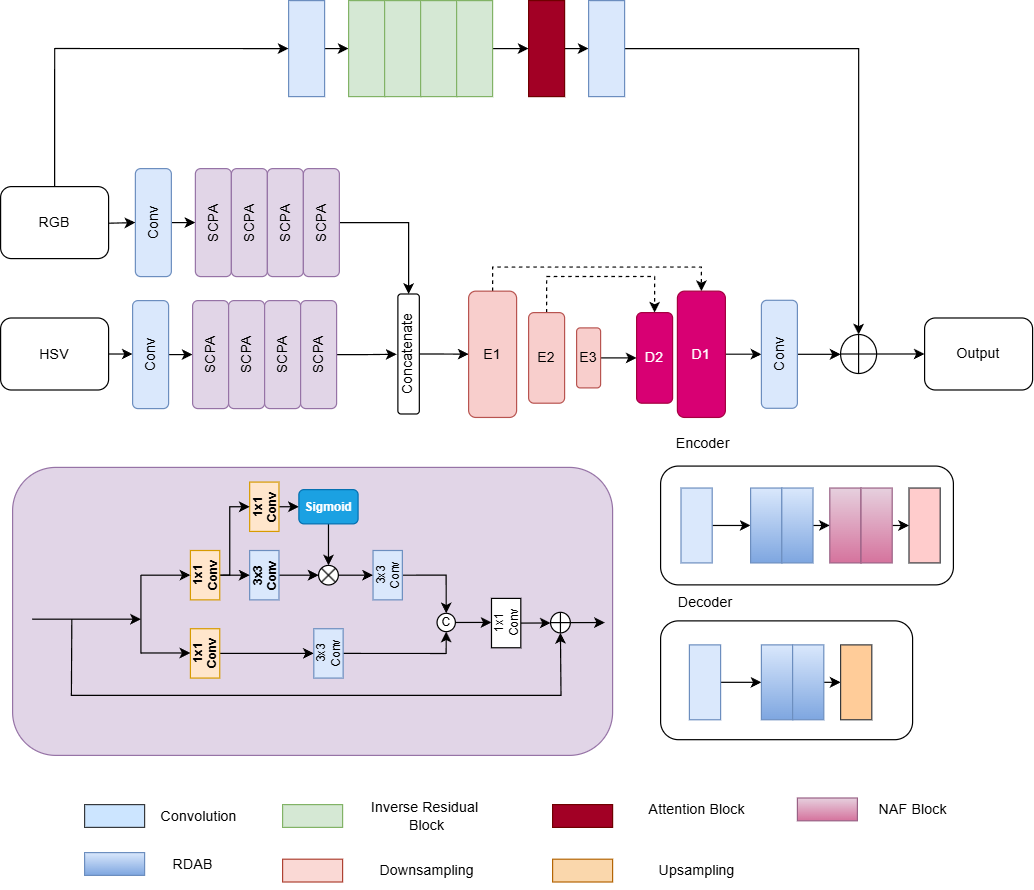}
    \caption{Overview of the Proposed Lightweight Self-Calibrated Pixel-Attentive Network model.}
    \label{fig:model_framework22}
\end{figure}
\begin{figure}  
	\centering
\includegraphics[width=1\linewidth]{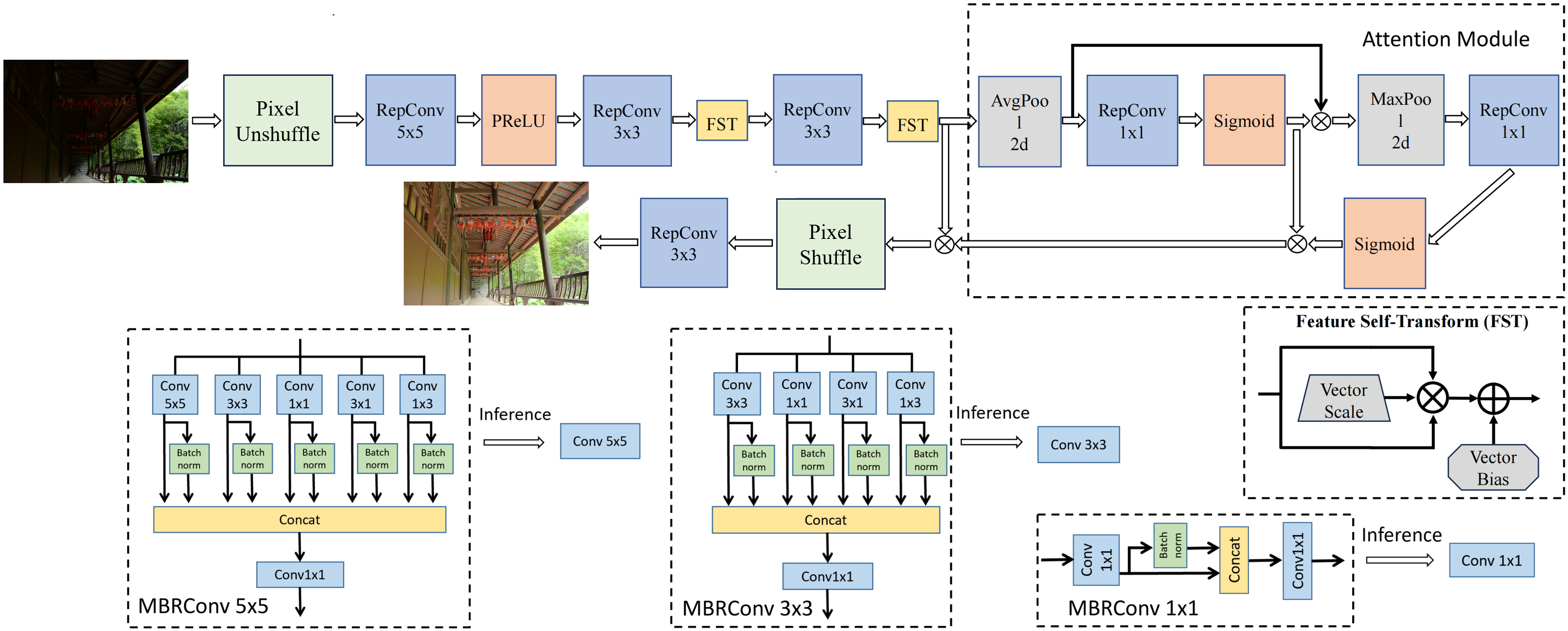}
	\caption{General method flow chart.}
	\label{AVC2}
\end{figure}
\begin{figure}  
	\centering
\includegraphics[width=1\linewidth]{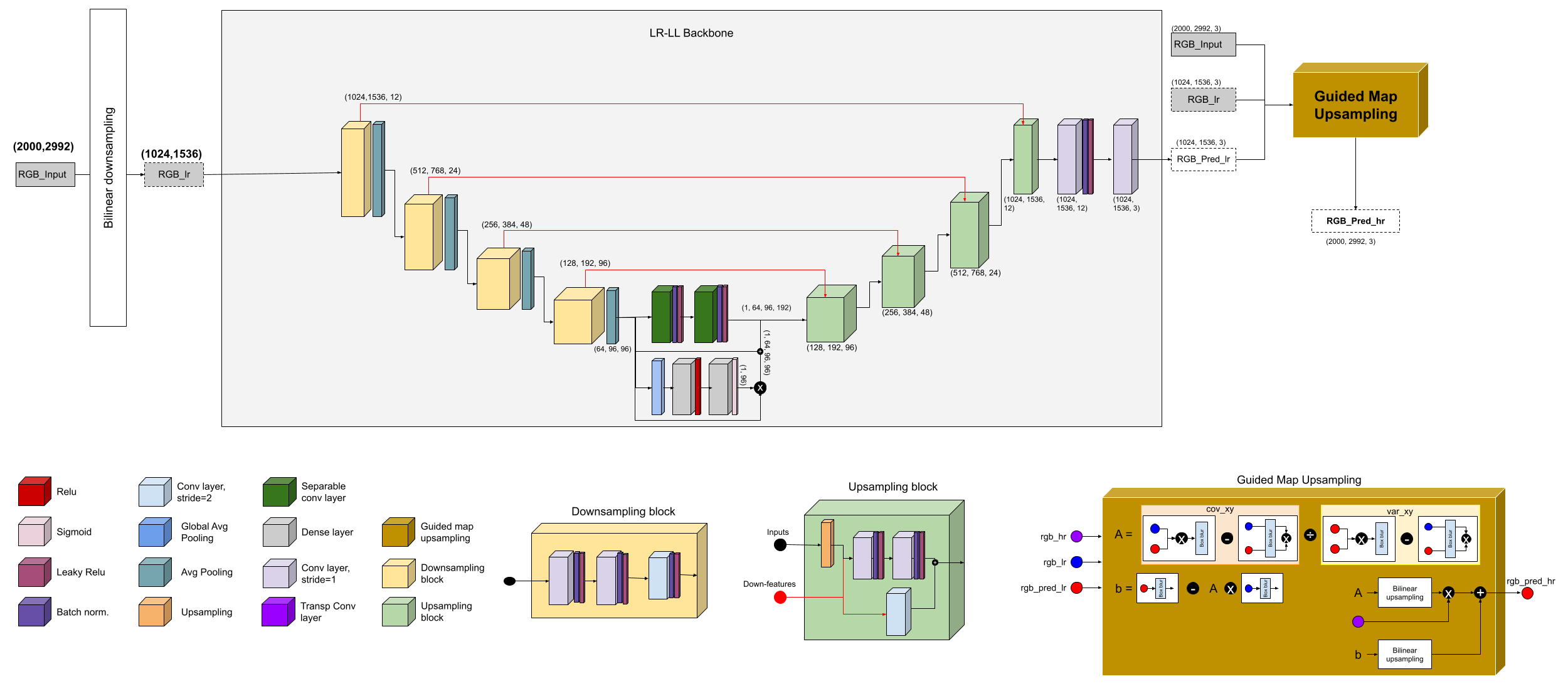}
	\caption{The high-level architecture of Team LR-LL.}
	\label{fig:LR-LL-Model}
\end{figure}
\begin{figure}
\centering
\includegraphics[width=23em]{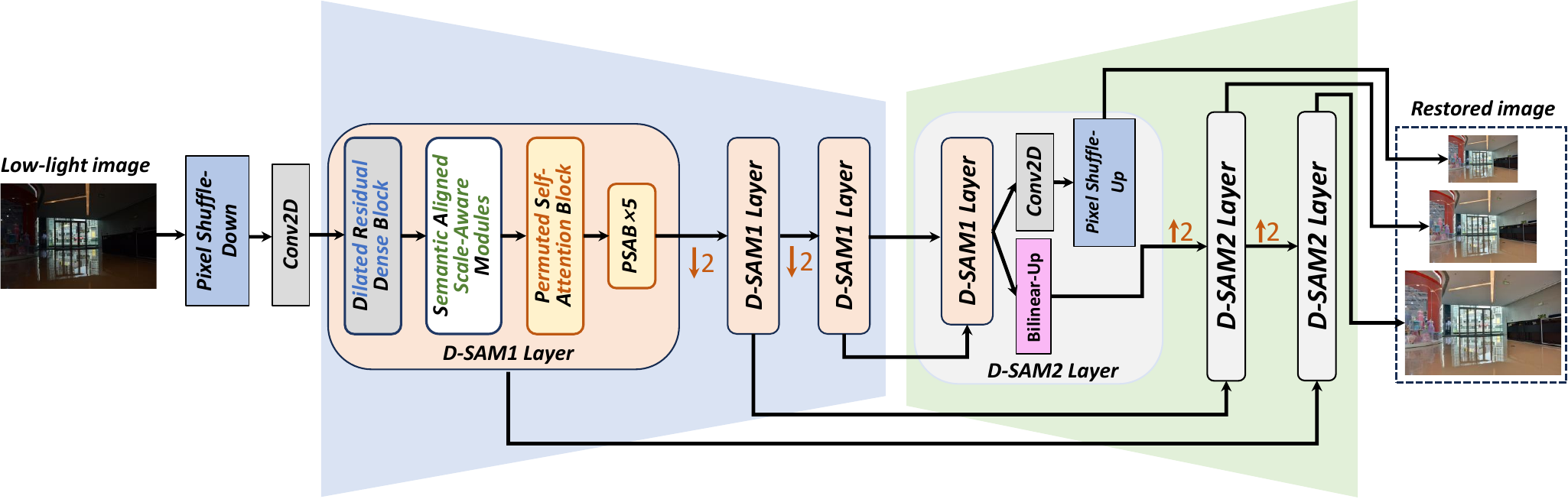}
\caption{{Overview of the proposed pipeline of Team X-L.}
\label{fig:llie}}
\end{figure}
\begin{figure}
\centering
\includegraphics[width=23em]{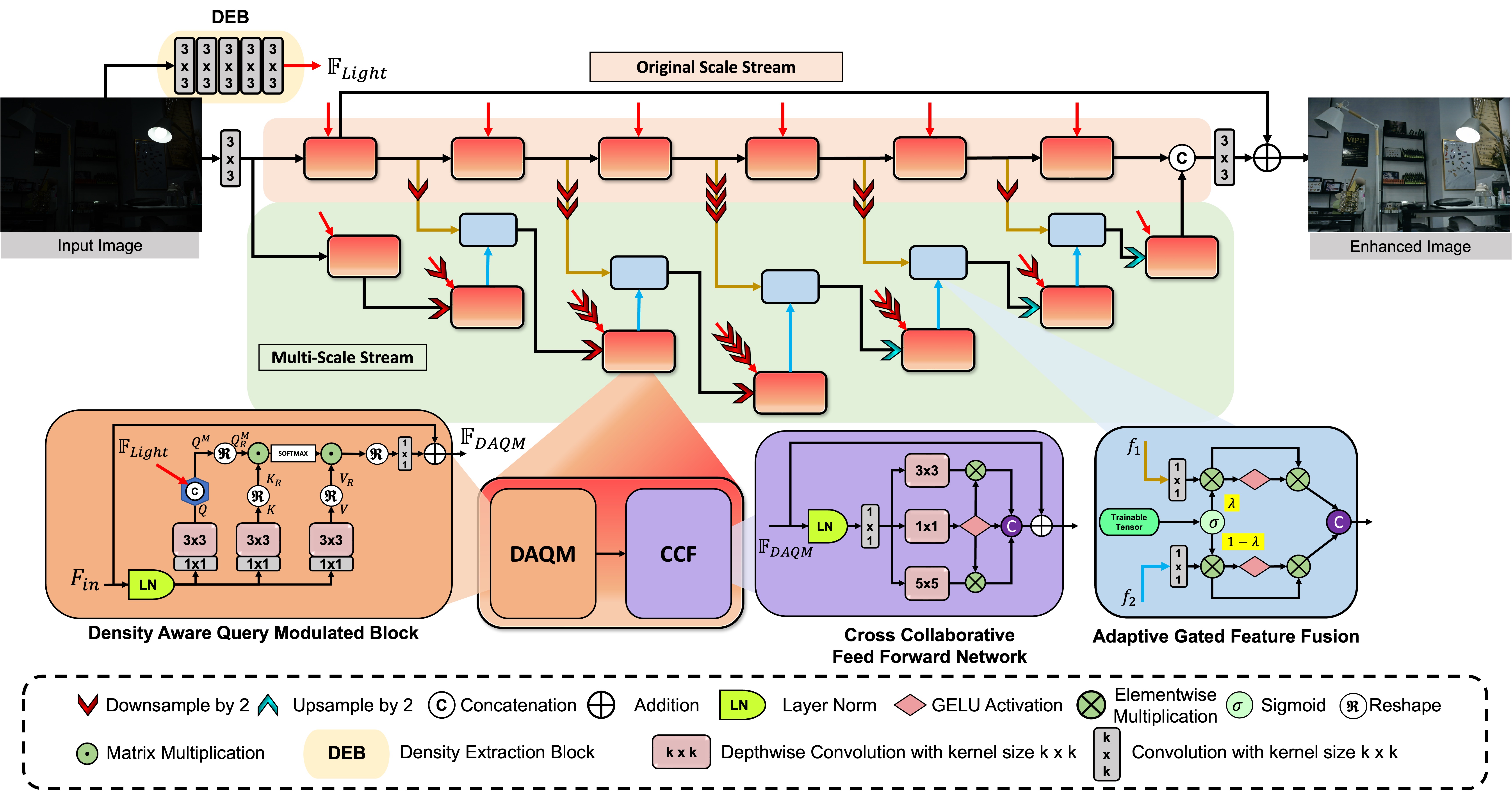}
\caption{{Overview of the IllumiNet for Lowlight Image Enhancement.}
\label{fig:TEAM_IITRPR}}
\end{figure}
\begin{figure}
    \centering
    \includegraphics[width=1\linewidth]{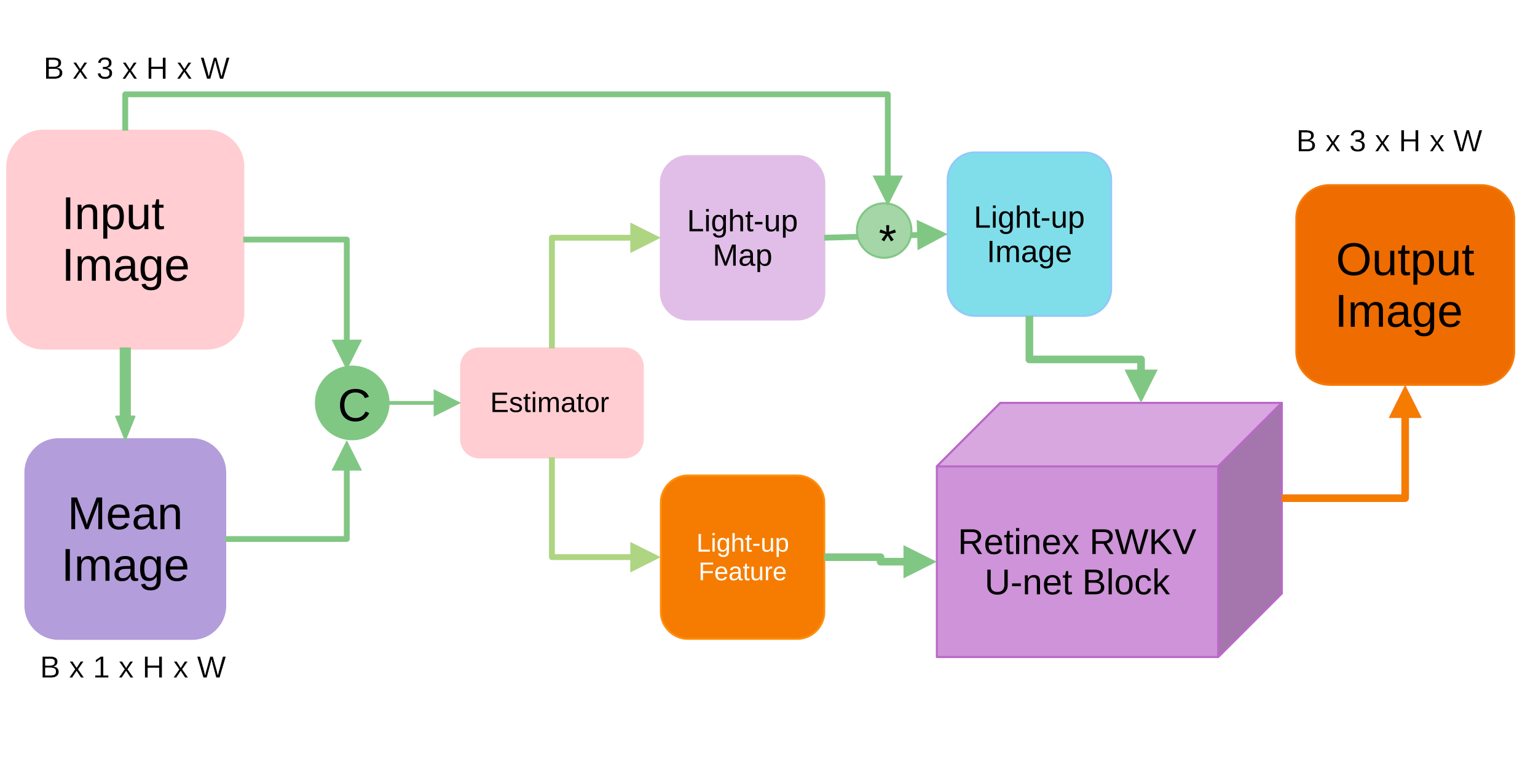}
    \caption{Overview of the RetinexRWKV of Team JHC-Info.}
    \label{fig:RetinexRWKV_backbone}
    \includegraphics[width=1\linewidth]{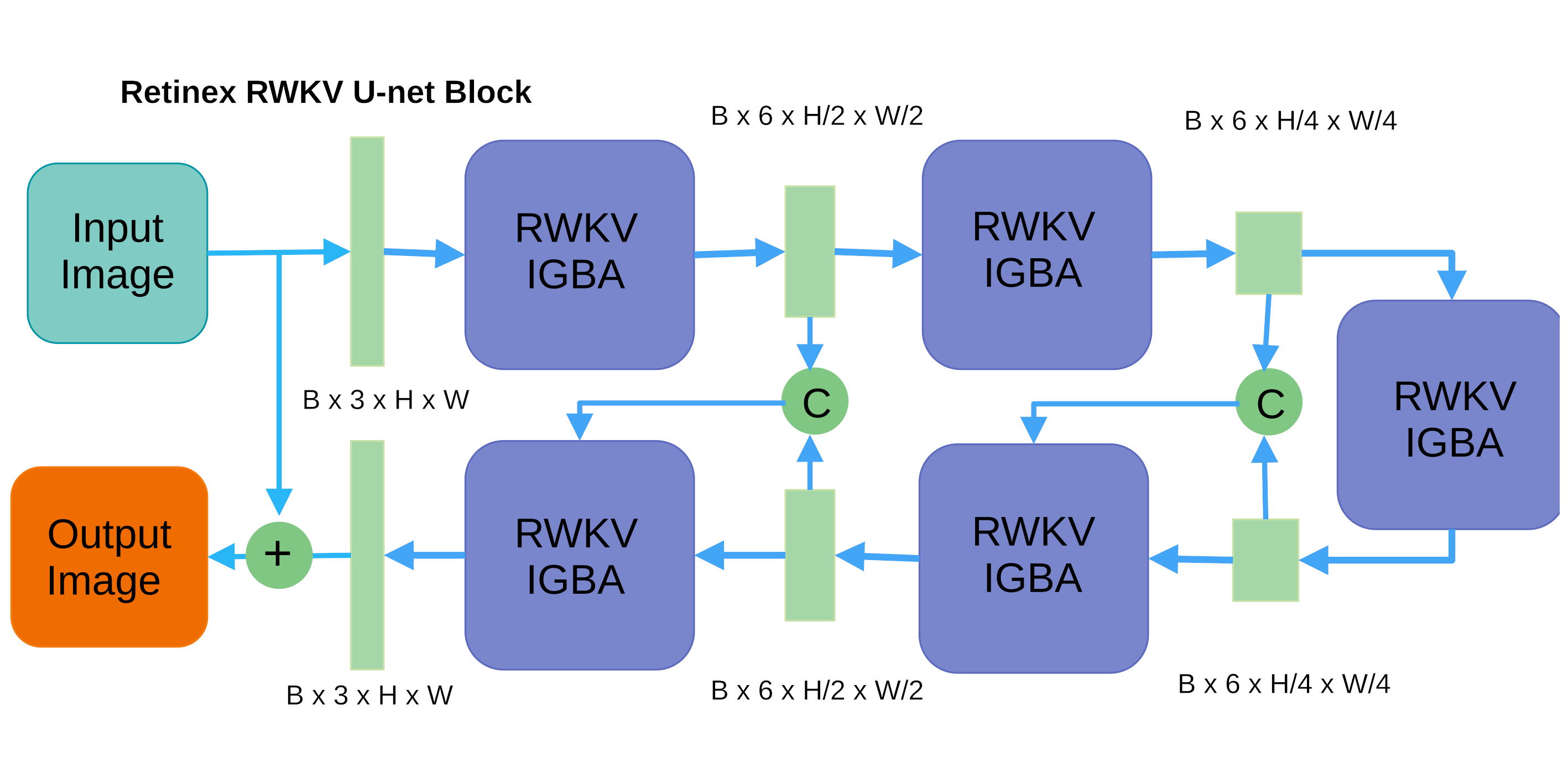}
    \caption{unetblock of the RetinexRWKV of Team JHC-Info.}
    \label{fig:RetinexRWKV_unetblock}
    \includegraphics[width=1\linewidth]{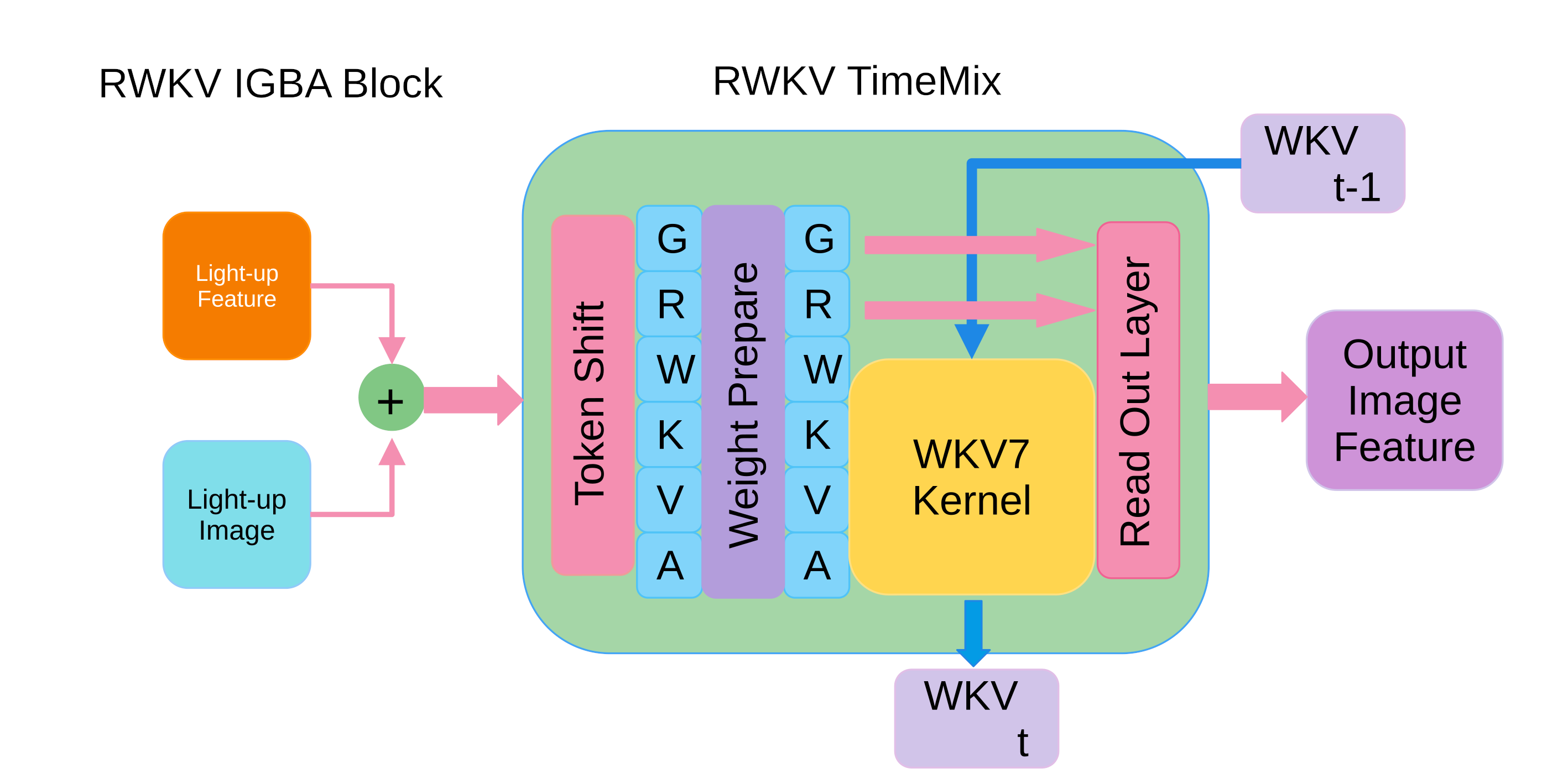}
    \caption{RWKV-v7 timemix of the RetinexRWKV of Team JHC-Info.}
    \label{fig:RetinexRWKV_timemix}
\end{figure}
\end{document}

%% file: final_results.tex
\begin{table*}[!t]
\footnotesize
\centering
\caption{Evaluation and Rankings in the NTIRE 2025 Low Light Image Enhancement Challenge. ``Rank"s indicate the respective standings of participants based on their performance in different metrics on the challenge's test dataset. ``Final Rank" represents a composite metric, derived from a weighted sum of 0.5, 0.5, 0.4, 0.2, respectively.}
\vspace{-3mm}
\begin{tabular}{cccccccccc}
\toprule
\textbf{Team}                & \textbf{PSNR} & \textbf{SSIM} & \textbf{LPIPS}  & \textbf{NIQE} & \textbf{Rank PSNR} & \textbf{Rank SSIM} & \textbf{Rank LPIPS}  & \textbf{Rank NIQE} &\textbf{Final Rank} \\ 
\midrule
NWPU-HVI  &26.24 &0.861 &0.128 &10.95 &2 &2 &7 &11 &1\\
Imagine  &26.35 &0.858 &0.133 &11.81 &1 &3 &9 &23 &2\\
pengpeng-yu & 25.85 & 0.858 & 0.134 & 11.29 & 4 & 3 & 11 & 16 & 3 \\
DAVIS-K  &25.14 &0.863 &0.127 &10.58 &14 &1 &6 &9 &4\\
SoloMan & 25.80 & 0.856 & 0.13 & 11.49 & 5 & 6& 8 & 19 & 5 \\
\rowcolor{gray!20} Smartdsp & 25.47 & 0.848 & 0.12 & 10.53 & 11 & 12 & 3 & 8 & 6 \\
Smart210  &26.15 &0.855 &0.137 &11.52 &3 &7 &14 &20 &7\\
WHU-MVP & 25.76 & 0.855 & 0.138 & 11.21 & 7 & 7 & 15 & 13 & 8\\
\rowcolor{gray!20} BUPTMM & 25.67 & 0.855 &0.137 &11.28 &8 &7 &14 &14 & 9 \\
NJUPT-IPR  &25.01 &0.848 &0.122 &10.15 &15 &12 &5 &3 &10\\
SYSU-FVL-T2  &25.65 &0.857 &0.135 &11.59 &10 &5 &12 &22 &11\\
KLETech-CEVI              & 25.66      &0.854      & 0.134        &11.55                  & 9                  &10 & 11 &21 &12                  \\
Ensemble-KNights  &25.77 &0.849 &0.139 &11.47 &6 &11 &16 &18 &13\\
MRT-LLIE             & 24.52      &0.833      & 0.117        &10.23                  & 19                  &18 & 2 &4 &14                  \\
\rowcolor{gray!20} SynLLIE & 24.01 &0.84 &0.117 &10.37 &22 &15 &2 &5 &15 \\
Cidaut AI             & 25.45      &0.839      & 0.144        &10.45                  & 12                  &16 & 17 &7 &16                  \\
Huabujianye & 25.15 &0.845 &0.157 & 11.17 &13 &14 &20 &12 & 17\\
\rowcolor{gray!20} no way no lay & 24.64 & 0.839 &0.154 &11.32 &17 &16 &18 &17 &18\\
Lux Themps             & 22.27      &0.822      & 0.122        &10.39                 & 27                  &21 &5 &6 &19                  \\
PSU\_team    &24.86 & 0.824 & 0.176 & 10.95 & 16 &20 &21 &10 &20\\
hfut-lvgroup    &24.54 &0.832 &0.157 &11.29 &18 &19 &20 &15 &21\\
ImageLab  & 23.87      &0.816      & 0.191        &9.68                  & 23                  &22 & 22 &1 &22                  \\
AVC2 & 24.02 &0.816 &0.196 &12.88 &21 &22 &23 &28 &23 \\
LR-LL  & 24.22      &0.816      & 0.236        &12.10                  & 20                  &22 & 27 &25 &24                  \\
X-L& 23.49 &0.803 &0.212 &12.86 &25 &26 &25 &27 & 25\\ 
Team\_IITRPR            & 23.50     &0.803     & 0.212        &12.86                 & 25                 &26 & 25 &27 &26                  \\
\rowcolor{gray!20} CV-SVNIT & 16.85 &0.565 &0.427 &12.29 &28 &28 &28 & 26&27 \\
\rowcolor{gray!20} JHC-Info & 23.38 &0.803 & 0.203 & 9.85 & 26 & 26 & 24 & 2 & - \\
\bottomrule
\end{tabular}
\label{tbl:ntire24_results}
\vspace{-2mm}
\end{table*}

%% file: team01_NWPU_HVI/main.tex
\subsection{NWPU-HVI}
\begin{figure}[!t]
	\centering
	\includegraphics[width=\linewidth]{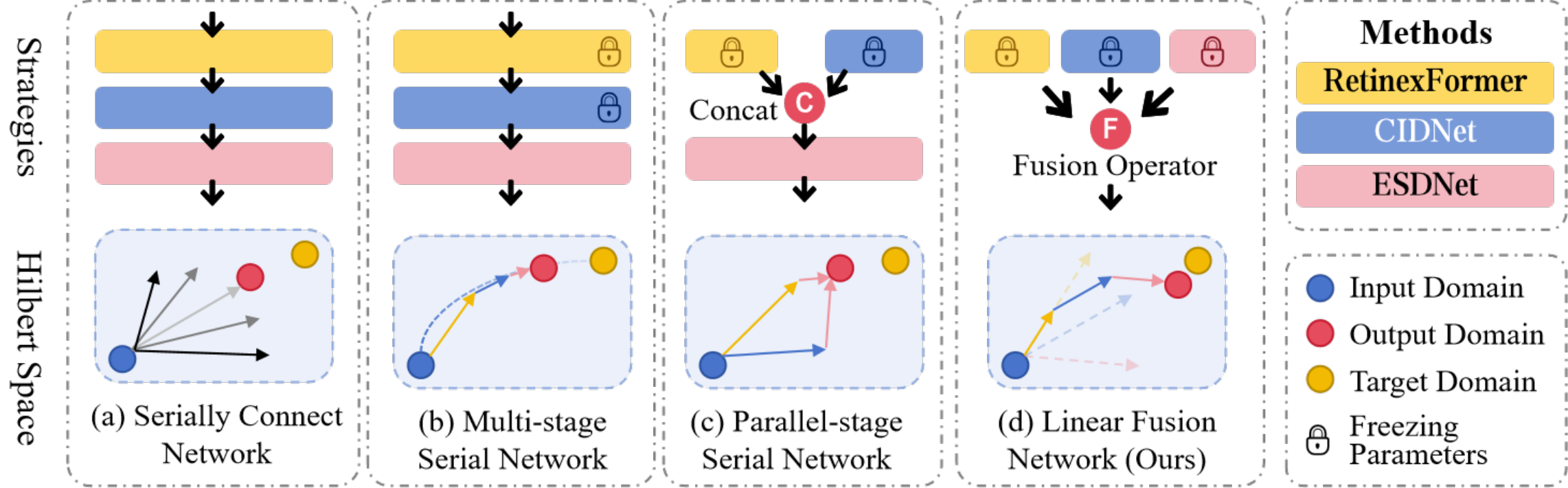}
	\vspace{-7mm}
	\caption{Architecture of 1st place solution.}
	\vspace{-3mm}
	\label{team1}
\end{figure}

\textbf{Description:}~We propose FusionNet, as shown in~\cref{team1}, a hybrid framework combining three complementary methods:~ESDNet \cite{yu2022towards} for local feature processing, Retinexformer \cite{retinexformer} for long-range dependencies, and CIDNet \cite{yan2025hvi} utilizing the HVI color space. FusionNet introduces four fusion strategies:~1) Serial network (single-stage, risks instability);
2) Multi-stage serial training (frozen parameters, slower convergence); 3) Parallel training followed by serial enhancement (inefficient); 4) Linear fusion with fully parallel execution with weights:
$\mathbf{I}_{HQ} = \sum_{i=1}^{n} k_i\mathtt{F}_i(\mathbf{I}_{LQ})$

Each method defines a unique mapping in Hilbert space, and optimal fusion maximizes projection in the target subspace, enhancing domain generalization.

\noindent\textbf{Implementation:}~The model is implemented in PyTorch with the Adam optimizer and cosine annealing schedule.~CIDNet/ESDNet/Retinexformer is trained on $1024^2$/$1600^2$/$2000^2$ patches for 90k/100k/180k iterations, respectively, all with a batch size 1. The training configuration is listed in \cref{tab:training_config}.

%% file: team02_Imagine/main.tex
\subsection{Imagine}
\begin{figure}
    \centering
    \includegraphics[width=\linewidth]{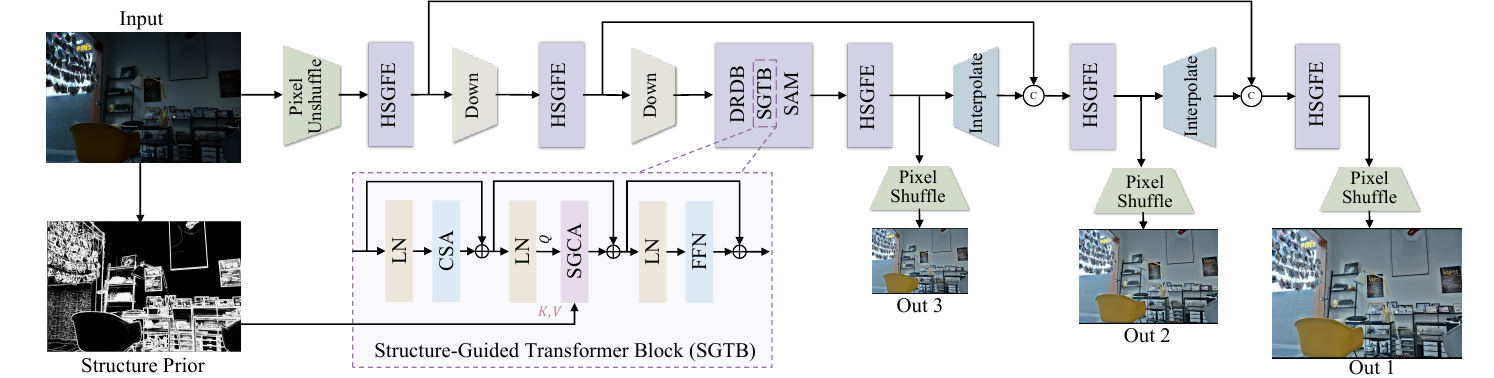}
        \vspace{-7mm}
    \caption{Architecture of 2nd place solution.}
    \vspace{-7mm}
    \label{team2}
\end{figure}

\noindent\textbf{Description:}~We propose SG-LLIE,~as shown in \cref{team2}, a multi-scale CNN-Transformer hybrid network based on UNet architecture.~The Hybrid Structure-Guided Feature Extractor (HSGFE) employs structural cues to preserve fine details.~Down-sampling is achieved with ``PixelUnshuffle'' and convolutional layers, while up-sampling uses ``PixelShuffle'' or ``Interpolate''.~Skip connections maintain spatial coherence.~The Color Invariant Convolution (CIConv) extracts illumination-invariant priors, and the Structure-Guided Transformer Block~(SGTB) modulates learning with Channel-wise Self-Attention (CSA), Structure-Guided Cross Attention (SGCA), and Feed-Forward Networks (FFN).~The model employs Dilated Residual Dense Blocks (DRDB) and Semantic-Aligned Scale-Aware Modules (SAM) for hybrid local and long-range feature learning. Training samples are classified by illumination levels, with adjustment factors applied. A self-ensemble strategy further enhances performance.~The total loss is a combination of combines Charbonnier loss~\cite{chan}, perceptual loss~\cite{vggloss}, and Multi-Scale SSIM loss~\cite{zhou2024glare} with a weighting of 1, 0.01 and 0.4, respectively.

\noindent\textbf{Implementation:}  
The model is trained on the NTIRE 2025 dataset with a patch size of $1600^2$, batch size of 1, initial learning rate of $2 \times 10^{-5}$, and 60,000 iterations with a cyclic cosine annealing schedule. The adjustment layer is trained with cross-entropy loss and softmax activation. The network is fine-tuned on the NTIRE 2024 and 2025 datasets, with a learning rate of $2 \times 10^{-7}$.

%% file: team03_pengpeng_yu/main.tex
\subsection{pengpeng-yu}
\noindent\textbf{Description:}~We adopted and extended the EDSNet~\cite{yu2022towards} implementation developed by the SYSU-FVL-T2 team in the NTIRE 2024 edition~\cite{liu2024ntire}. Our method builds upon an encoder-decoder architecture with three feature scales and skip connections. At each scale, Semantic-Aligned Scale-Aware Modules are stacked to enhance the model's ability to process multi-scale features. The loss function integrates the Charbonnier loss~\cite{chan} and the perceptual VGG19 loss~\cite{vggloss}. To mitigate semantic degradation during downsampling, we enable anti-aliasing in the downsampling operations within the SAM modules and the perceptual loss computation.~For inference, we adopt a self-ensembling strategy, applying geometric transformations to the input and averaging the corresponding predictions.

\noindent\textbf{Implementation:}
We train the model for 150,000 iterations using the Adam optimizer on an NVIDIA RTX 4090 GPU, starting from the pretrained SYSU-FVL-T2 model weights with a resolution of 1600 and an initial learning rate of 0.0002, which is scheduled using cyclic cosine annealing. Inference is conducted at the original resolution.

%% file: team04_DAVIS_K/main.tex
\subsection{DAVIS-K}
\noindent\textbf{Description:}~We propose a two-stage U-shaped Transformer, as illustrated in~\cref{fig:fig4-1}.~The model comprises an \textit{Enhancement Module} and a \textit{Refinement Module}. The Enhancement Module contains two branches. The first branch adopts Restormer~\cite{zamir2022restormer}, a hierarchical U-Net with skip connections and Transformer blocks that increase in depth from top to bottom. It employs Multi-Dconv Head Transposed Attention for efficient channel-wise feature interaction. Input images are downsampled before processing and upsampled via PixelShuffle. 

The second branch is designed for transfer learning. It uses the first three layers of a pre-trained ConvNeXt~\cite{liu2022convnet} as the encoder to leverage prior knowledge, followed by a lightweight decoder~\cite{zhou2023breaking} for gradual upsampling. The Refinement Module, sharing the same architecture as the first branch, further enhances color and texture details.

\noindent\textbf{Implementation:} The model is trained using the Adam optimizer ($\beta_1=0.9$, $\beta_2=0.999$) for 600K iterations with a cosine-annealed learning rate from $2\times10^{-4}$ to $1\times10^{-6}$. We use a batch size of 8 and randomly crop $384\times384$ patches with rotation and flip augmentations. Training is conducted in PyTorch on an RTX 4090 GPU. This work is supported by the Technology Development Program (RS-2024-00469833), funded by the Ministry of SMEs and Startups (MSS, Korea).

%% file: team05_SoloMan/main.tex
\subsection{SoloMan}
\textbf{Description:} We propose \textbf{ESDNet-Twins}, a twin-network architecture employing a complementary ensemble strategy for low-light image enhancement. The first network, \textit{Twin1}, follows the multi-scale design of ESDNet~\cite{yu2022towards}, which already delivers strong performance. However, it struggles in extremely over-/under-exposed regions and lacks in detail preservation and edge sharpness.

To address these limitations while retaining its strengths, we introduce \textit{Twin2}—a lightweight counterpart with improved DB, RDB, and SAM blocks, and a novel multi-scale \textbf{Feature Full Connect Block} between the encoder and decoder. This design enables Twin2 to handle higher-resolution images more efficiently. Instead of merging both networks into a single model—which could lead to overfitting due to the small dataset—we train them independently and apply an ensemble strategy, as shown in Fig.~\ref{fig:enter-label_pipnline}.~We optimize the model using a multi-VGG perceptual loss:
\vspace{-2mm}
\begin{equation}
	\mathcal{L} = \mathcal{L}_{\text{VGG}} + \mathcal{L}_{\text{CB}},
	\vspace{-2mm}
\end{equation}
where $\mathcal{L}_{\text{VGG}}$ denotes perceptual loss using a pretrained VGG16, and $\mathcal{L}_{\text{CB}}$ is the Charbonnier loss.

\noindent\textbf{Implementation:}~ESDNet-Twins is implemented in PyTorch and trained on an NVIDIA A6000 (48GB). \textit{Twin1} is trained from scratch for 150K iterations using Adam. Training is staged with decreasing batch sizes \{8, 4, 4, 2, 2, 1\} and corresponding patch sizes \{720, 1024, 1024, 1280, 1600\}.~\textit{Twin2}, being more efficient, follows the same setup but with larger patches \{720, 1024, 1024, 1600, 1800\}. Both models use an initial learning rate of $2\times10^{-4}$, scheduled with cyclic cosine annealing.

%% file: team06_Smartdsp/main.tex
\subsection{Smartdsp  (Excluded from Report)}

%% file: team07_Smart210/main.tex
\subsection{Smart210}

\textbf{Description:} Since PSNR and SSIM are the primary evaluation metrics, we adopt ESDNet~\cite{yu2022towards}, a strong SOTA baseline, as our foundation. ESDNet employs an encoder-decoder structure with Dilated Residual Dense Blocks for feature extraction and Semantic-Aligned Scale-Aware Modules (SAM) for multi-scale fusion. As shown in \cref{fig7-1}, we replace the Dilated Residual Dense Block with the SimPFblock~\cite{wan2024psc}, which is based on the NAFBlock~\cite{chen2022simple} and incorporates the parameter-free attention mechanism SimAM~\cite{yang2021simam}, illustrated in \cref{fig7-2}. SimPFblock reduces multiplication operations—beneficial for low-light inputs where low pixel values may cause vanishing gradients—making it well-suited for this task. Experiments confirm that SimPFblock improves performance in PSNR and SSIM.

\noindent\textbf{Implementation:} Our method is implemented in Python 3.8 using PyTorch. Following the progressive training strategy in SYSU-FVL-T2~\cite{liu2024ntire}, we train for 250K iterations using the Adam optimizer ($\beta_1=0.99$, $\beta_2=0.999$) with an initial learning rate of 0.0002, decayed via cyclic cosine annealing. Final predictions are generated by linearly combining outputs from checkpoints at 150K, 200K, and 205K iterations.~We also adopt the self-ensemble strategy from Retinexformer~\cite{retinexformer}, which consistently boosts PSNR.

%% file: team08_WHU_MVP/main.tex
\subsection{WHU-MVP}
\noindent
\noindent\textbf{Description:}~We adopt ESDNet~\cite{yu2022towards} as the backbone for LLIE.~Given the high-resolution nature of low-light images, computing global attention directly can be computationally expensive. To address this, inspired by Transformer-based approaches~\cite{zamir2022restormer}, we introduce a parallel restoration branch operating on $4\times$ downsampled inputs to perform coarse enhancement. The output is then refined by ESDNet at the original resolution. This coarse-to-fine strategy enhances adaptability to high-resolution inputs and improves overall restoration quality. Extensive experiments validate the effectiveness of this design.

\noindent\textbf{Implementation:}~Training is performed exclusively on the official competition dataset using four NVIDIA RTX 4090 GPUs. A progressive training scheme is employed with increasing patch sizes \{1024, 1280, 1280, 1600, 1920\} and corresponding batch sizes \{4, 4, 2, 2, 1\}, over \{46K, 32K, 24K, 18K, 30K\} iterations, respectively. The initial learning rate is $2 \times 10^{-4}$, scheduled via cyclic cosine annealing. For inference, images under $3000\times3000$ are processed directly, while larger images are split, processed in segments, and reassembled to reduce GPU memory usage.

%% file: team09_BUPTMM/main.tex
\subsection{BUPTMM (Excluded from Report)}

%% file: team10_NJUPTIPR/main.tex
\subsection{NJUPT-IPR}
\textbf{Description:} We adopt an enhanced version of our previous work IIAG-CoFlow~\cite{10845035} for this competition. IIAG-CoFlow is a normalizing flow-based method for low-light image enhancement, comprising a conditional generator and a complete flow module. IIAG is a U-shaped Transformer network built with IIZAT (inter-/intra-channel and zeromap attention Transformer) for downsampling and IIAT (inter-/intra-channel attention Transformer) for upsampling. IIAT models inter- and intra-channel attention independently, while IIZAT further integrates zeromap attention in parallel. CoFlow introduces three novel invertible transformations—linear injector, conditional linear coupling, and unconditional linear coupling—guided by a cross-attention network to model affine transformations conditioned on features. During training, CoFlow takes $I_H$, $Ft1$, $Ft2$, and $Ft3$ as inputs to produce latent variables $z = \Phi(I_H; I_L)$, which are mapped to $\mathcal{N}(0,1)$. In inference, $z$ and feature maps generate the enhanced image $I_E$ from the low-light input $I_L$.

\noindent\textbf{Implementation:} The method is implemented in PyTorch and trained with the Adam optimizer for $150{,}000$ iterations. The learning rate starts at $2 \times 10^{-4}$ and is halved at 25\%, 50\%, 70\%, and 80\% of training.~Training uses $896 \times 896$ random crops with a batch size of 1. At test time, $2000 \times 3000$ images are processed directly, while $4000 \times 6000$ images are split into four tiles and merged post-enhancement. This work is supported by the National Natural Science Foundation of China (Grant 62272240).

%% file: team11_SYSU_FVL_T2/main.tex
\subsection{SYSU-FVL-T2}
\noindent\textbf{Description:}~We propose a low-light image enhancement method based on ESDNet-L~\cite{yu2022towards}, as shown in~\cref{fig_esdnet}. The model adopts an encoder-decoder architecture with three feature scales, connected via skip connections.~Multi-scale features are generated using Lanczos3 interpolation. At each scale, two stacked Semantic-Aligned Scale-Aware Modules (SAM) are used to enhance the model’s ability to handle scale variations.~Each SAM integrates a pyramid context extraction module and a cross-scale dynamic fusion module for selective multi-scale fusion. The total loss $L_{total}$ is defined for the outputs at three scales with a combination of Charbonnier loss~\cite{chan}, perceptual loss~\cite{vggloss}, and Multi-Scale SSIM loss~\cite{zhou2024glare}, and the color loss~\cite{kind}, with a weighting of 1, 0.04, 1 and 1, respectively.

\noindent\textbf{Implementation:}~The method is implemented in Python 3.8 and trained on an NVIDIA RTX A6000 (49GB). Following the progressive training strategy of MIRNet-v2~\cite{mirv2}, we train the model from scratch for 156,000 iterations using the Adam optimizer~\cite{adam}. Initially, we use a batch size of 8 with $720\times720$ patches, gradually adjusting to batch sizes of 4, 4, 2, 2, 1, and 2, and patch sizes of 1024, 1024, 1280, 1280, 1600, and 1440 at respective iteration stages (46k, 32k, 24k, 18k, 18k, and 6k). The learning rate starts at $2\times10^{-4}$ and follows a cyclic cosine annealing schedule~\cite{sgdr}. During inference, the full-resolution image is processed directly with a batch size of 1.

%% file: team12_KLETech_CEVI/main.tex
\subsection{KLETech-CEVI}
\noindent\textbf{Description:}~We propose ESDNet+, as shown in \cref{fig:esdnet+}, an enhanced version of ESDNet \cite{yu2022towards}, for efficient low-light enhancement of ultra-high-definition (4K) images. The method utilizes a single-stage pipeline with a pre-processing head, encoder-decoder architecture, and intermediate supervision. Key components like Dilated Residual Dense Blocks (DRDB) and Semantic-Aligned Scale-Aware Modules (SAM) are retained, with the introduction of a novel Low-Light Enhanced Perceptual Loss. The pipeline begins by downsampling the input image and extracting features with a 5$\times$5 depth-wise convolution. These features are processed through a three-level encoder with DRDB and SAM for multi-scale feature fusion. The encoder’s output is upsampled in the decoder, with skip connections to preserve high-resolution details.~The final output is a fully enhanced 4K image, with intermediate outputs supervised during training. The loss function combines perceptual, luminance, and edge-preserving losses to optimize performance in low-light conditions:
\vspace{-2mm}
\begin{equation}
	L_{\text{ESDNet+}} = \alpha \cdot L_{\text{VGG}} + \beta \cdot L_{\text{Luminance}} + \gamma \cdot L_{\text{Edge}}.
	\vspace{-2mm}
\end{equation}
\noindent\textbf{Implementation:}~Participants did not provide details.

%% file: team13_Ensemble_KNights/main.tex
\subsection{Ensemble-KNights}
\noindent\textbf{Description:}~We propose the \textbf{EN}semble~\textbf{B}ayesian ~\textbf{E}nhancement~\textbf{M}odel~(\textbf{EN-BEM}), an enhancement of BEM~\cite{huang2025bayesian} that integrates Transformer and Mamba architectures. EN-BEM leverages BEM's two-stage approach, where a Bayesian Neural Network (BNN) models one-to-many mappings in the first stage, and a Deterministic Neural Network (DNN) refines image details in the second stage. In EN-BEM, the backbone is replaced with either a Transformer or Mamba architecture, with outputs combined through internal ensembling. This ensemble method balances computational efficiency,~noise suppression, and detail restoration, while addressing the one-to-many mapping issue in low-light enhancement. The probabilistic nature of the BNN enables EN-BEM to capture data uncertainty, making it robust in dynamic low-light conditions. The overall framework is shown in \cref{fig:enbem}.

\noindent\textbf{Implementation:} The models are implemented in PyTorch and trained on the provided dataset without external data. Training and testing are performed on a single RTX 4090 GPU. The Adam optimizer is used with an initial learning rate of $2 \times 10^{-4}$, decaying to $10^{-6}$ using a cosine annealing schedule. The first-stage model is trained for 300K iterations on $1792 \times 1792$ inputs, and the second-stage model for 150K iterations on $496 \times 496$ inputs, with a batch size of 8. During inference, images are processed at full resolution, with a batch size of 1. A self-ensemble technique is applied during testing.

%% file: team14_MRT_LLIE/main.tex
\subsection{MRT-LLIE}
\noindent\textbf{Description:}~We propose \textbf{MRT}, a novel Transformer network leveraging a new encoder-decoder scheme called the \textbf{Multi-scale Entanglement Scheme}. Inspired by~\cite{liu2024ntire} (Sec. 4.16), this scheme is tailored for Transformers to learn enhanced multiscale feature representations. Additionally, we introduce a \textbf{Residual Multi-headed Self-Attention} mechanism to preserve details across network stages. The \textbf{Multi-stage Squeeze \& Excite Fusion Block}~\cite{lytnet2024} is incorporated in the post-attention step for improved feature extraction. The design of \textbf{MRT} is shown in \cref{fig:mrt_framework}.

\noindent\textbf{Implementation:}~MRT is implemented in PyTorch, trained on the NTIRE25 dataset. The model is optimized using the Adam optimizer for 150k iterations, with an initial learning rate of 2$e$-4, decaying via Cosine Annealing. Each iteration uses a batch of two $704 \times 704$ randomly-cropped image patches with data augmentation (random flipping/rotation). We employ a hybrid loss function that captures pixel-level, multi-scale, and perceptual differences.~Testing is conducted via standard inference, except for $4000 \times 6000$ images, which are split into four $4000 \times 1500$ images using pixel interleaving to manage resource constraints.

%% file: team15_SynLLIE/main.tex
\subsection{SynLLIE (Excluded from Report)}

%% file: team16_CidautAI/main.tex
\subsection{Cidaut AI}
\noindent\textbf{Description:}~We propose two original models: \textbf{FLOL}~\cite{benito2025flol} and \textbf{DarkIR}~\cite{feijoo2024darkir}, both utilizing Fourier frequency information and the NAFBlock~\cite{chen2022simple} architecture. The architecture, as shown in~\cref{fig:method_cidautai}, consists of two stages in each network: (1) an illumination stage that enhances the image to the optimal lightness, and (2) a Denoiser stage (FLOL) or Deblur stage (DarkIR) that refines the enhanced image by removing noise, blur, and imperfections from the first stage. For FLOL, we use the Semantic-Aligned Scale-Aware Modules (SAM)~\cite{yu2022towards}~loss, combining perceptual and distortion terms across multiple crop sizes to improve performance. The loss is defined as:
\vspace{-2mm}
\begin{equation}
    \mathcal{L} = \sum\nolimits_{i=1}^{3} \left( \mathcal{L}_1 + \mathcal{L}_{inter} + \lambda \mathcal{L}_{LPIPS} \right),
    \vspace{-2mm}
\end{equation}
where $\lambda = 0.1$, $\mathcal{L}_1$ is the $L_1$ loss (MAE), $\mathcal{L}_{inter}$ is the $L_1$ loss for the intermediate image, $i$ is the hierarchical scale, and $\mathcal{L}_{LPIPS}$ is the perceptual loss from the VGG19 model~\cite{vgg19}. For DarkIR, the multisize sum loss is not implemented. Additional optimization details are provided in the respective papers~\cite{benito2025flol, feijoo2024darkir}.

\noindent\textbf{Implementation:}~Both models are implemented in PyTorch. For DarkIR, we use the Adam optimizer with weight decay $1\times10^{-3}$ and a learning rate of $1\times10^{-3}$, following a Cosine Annealing schedule down to $1\times10^{-7}$. Training consists of 5 stages with epoch lengths [250, 150, 100, 50, 50], varying crop sizes [384, 720, 1024, 1280, 1280] and batch sizes [24, 8, 4, 4, 4], using 4 H100 GPUs for approximately 6 hours. For FLOL, the Adam optimizer is used with a learning rate of $2\times10^{-4}$, also following a Cosine Annealing schedule.~The training comprises 9 stages with crop sizes [720, 1024, 1024, 1280, 1280, 1600, 2000, 2200, 2400] and batch sizes [8, 4, 4, 2, 2, 1, 1, 1, 1], trained on a single NVIDIA GeForce RTX 4090 GPU for approximately 72 hours across 2000 epochs. Both models utilize random square crops ($H=W$) and random vertical and horizontal flips for data augmentation.

%% file: team17_Huabujianye/main.tex
\subsection{D-RetinexMix}
\noindent\textbf{Description:}~We propose an efficient multiscale network architecture that adapts DiffLL-based~\cite{jiang2023low} generated results to match the illumination conditions of the dataset. The enhanced outputs from the RetinexFormer~\cite{retinexformer} pre-trained model are fused using an objective evaluation metric to produce high-quality results. The entire process is implemented as an end-to-end training framework. Specifically, during training, DiffLL generates pre-enhanced results from low-light images in the training set. We then construct a U-shaped network with dual branches: Vmamba and convolution blocks.~The pre-enhanced results and the original low-light images are concatenated along the channel dimension and fed into this network to generate the second enhanced output.~Finally, the second enhanced results are evaluated against the outputs from the RetinexFormer pre-trained model using an objective metric to select the highest-quality images as the final output.

\noindent\textbf{Implementation:}~Experiments are conducted on a single NVIDIA GeForce RTX 3090 GPU with 24GB of memory, training for 50k iterations with random horizontal and vertical flipping.~The Adam optimizer is used with a learning rate of $4 \times 10^{-4}$, a patch size of $512 \times 512$, and a batch size of 8. The results from DiffLL and RetinexFormer are obtained using the pre-trained weights.

%% file: team18_no_way_no_lay/main.tex
\subsection{No Way No Lay (retimixformer)}
\noindent\textbf{Description:}~ we proposed the Improved Transformer Architecture Based on RetinexFormer with Quaternion Illumination Estimation model:retimixformer
To enhance illumination modeling capabilities while preserving detail and suppressing noise, we propose an improved Transformer-based architecture grounded in RetinexFormer~\cite{retinexformer}. Specifically, we introduce a novel quaternion illumination estimation module  to capture more expressive and physically consistent illumination representations. By encoding illumination conditions as quaternion-valued signals, allows the model to better disentangle lighting variations across spatial dimensions. This modification significantly improves the model's ability to perform robust low-light enhancement under complex and non-uniform illumination scenarios.

\noindent\textbf{Implementation:}~All models are trained on the NTIRE 2025 Low-Light Image Enhancement training dataset using two NVIDIA RTX A6000 GPUs (each with 48GB memory). The training process lasts for 48 hours, corresponding to approximately 30k iterations. We employ the AdamW optimizer with an initial learning rate of $1 \times 10^{-5}$ and a batch size of 32. Input images are uniformly cropped into $512 \times 512$ patches. To improve generalization, standard data augmentation techniques including random horizontal/vertical flipping and random rotation are applied during training.

%% file: team19_Lux_Themps/main.tex
\subsection{Lux Themps}
\noindent\textbf{Description:}~Incorporating semantic information from MobileNetV3~\cite{MobilNetV3}, our method, \textbf{SADe-ViT}, adapts illumination by distinguishing regions and objects.~The architecture follows a U-shaped encoder-decoder ViT design (see \cref{fig:sade_framework}), where Transformer blocks integrate segmentation maps into attention mechanisms, enhancing feature representation and illumination adjustment.~A multi-headed self-attention mechanism highlights key regions, followed by element-wise multiplication with spatially adapted semantic features to ensure dimensional alignment.~To improve computational efficiency, CNNs and fully connected layers in the FFN are replaced by Depthwise Separable Convolutions, as in~\cite{depthlux_mdpi, depthlux_isetc}, resulting in a model with only 0.57M parameters. The FFN output (\cref{eq:depthffn}) is normalized using Layer Normalization (LN).
\vspace{-2mm}
\begin{equation}
\label{eq:depthffn}
\mathbf{F}'_{\text{in}} = \text{DSC}^2 \sum\nolimits_{i=1}^{2} \text{DSC}^2 (\mathbf{F}_{\text{in}}), \quad \mathbf{F}'_{\text{in}} \in \mathbb{R}^{H \times W \times C}.
\vspace{-2mm}
\end{equation}

A hybrid loss function evaluates multiple aspects of the generated images:
\vspace{-2mm}
\begin{equation}
\mathcal{L} = \alpha \mathcal{L}_{2} + \beta \mathcal{L}_{perc} + \gamma \mathcal{L}_{SSIM},
\vspace{-2mm}
\end{equation}
where $\mathcal{L}_{2}$ is the mean squared error, $\mathcal{L}_{perc}$ is the perceptual loss using VGG-19~\cite{vggloss}, and $\mathcal{L}_{SSIM}$~\cite{ssim} is based on the Multi-Scale Structural Similarity Index.

\noindent\textbf{Implementation:}~Implemented in PyTorch, training is optimized using the Adam optimizer with a cosine annealing learning rate schedule, starting at $2 \times 10^{-4}$, increasing to $3 \times 10^{-4}$, and gradually decaying to $1 \times 10^{-6}$. We train for 300k iterations on $256 \times 256$ augmented patches from the LOL-v2 real~\cite{lol-v2} dataset, focusing on severe light-deficient images. For testing phase, Evaluation is performed on the competition's test dataset, including images of sizes $2000 \times 2992$ and $4000 \times 6000$. For the larger images, we apply a tile-based enhancement, splitting them into four tiles, enhancing each, and then reconstructing the full image.

%% file: team20_PSU_TEAM/main.tex
\subsection{PSU\_team}
\noindent\textbf{Description:}~We introduce \textbf{OptiMalDiff} that reformulates image denoising as an optimal transport problem.~The approach models the transition from noisy to clean images using a Schrödinger Bridge-based diffusion process. As shown in \cref{fig:method_psu}, the architecture comprises: 1) a hierarchical Swin Transformer backbone for efficient extraction of local and global features; 2) a Schrödinger Bridge Diffusion Module for learning forward and reverse stochastic mappings, and (3) a Multi-Scale Refinement Network (MRefNet) for progressively refining image details. Additionally, a PatchGAN discriminator is integrated for adversarial training to enhance realism.

\noindent\textbf{Implementation:}~The model is trained from scratch using the Low Light Image Enhancement dataset, without pre-trained weights.~We jointly optimize all modules with a composite loss function, combining diffusion loss, Sinkhorn-based optimal transport loss, multi-scale SSIM and $L_1$~losses, and an adversarial loss.~Training is conducted over 300 epochs with a batch size of 8, totaling 35,500 iterations per epoch.


%% file: team21_hfut_lvgroup/main.tex
\subsection{hfut-lvgroup}
\noindent\textbf{Description:}  
Our approach integrates Retinexformer \cite{retinexformer} with a U-Net variant \cite{feng2022mipi} using a simple yet effective model fusion strategy. Both models independently process the low-light image, with the final output obtained by weighted averaging. Retinexformer, grounded in Retinex theory and enhanced by the Transformer architecture, excels at global modeling, making it effective for complex lighting conditions. The U-Net variant combines the full-resolution features of FRC-Net \cite{zhang2023frc} with classic U-Net layers, learning both spatial structure and semantic information through stacked residual blocks at various scales.

\noindent\textbf{Implementation:}  
The models were trained separately using the NTIRE 2025 challenge dataset:
\begin{itemize}
    \item Retinexformer:~Trained on two NVIDIA RTX 4090 GPUs with a batch size of 1 for 150,000 iterations. The learning rate started at 1$e$-4 for the first 80,000 iterations and decreased to 1$e$-5 for the remaining iterations.
    \item U-Net Variant:~Trained on a single NVIDIA RTX 4090 GPU for 1,000 epochs, with a learning rate starting at 1$e$-4 and gradually reduced to 1$e$-5 for the first 500 epochs, then maintained at 2$e$-5 for the next 500 epochs.
\end{itemize}
Both models used the Adam optimizer and L1 loss, with random cropping of images into $1400\times1400$ patches to improve learning efficiency.

%% file: team22_ImageLab/main.tex
\subsection{ImageLab}
\noindent\textbf{Description:}~The Lightweight Self-Calibrated Pixel-Attentive Network for Low-Light Image Enhancement (LLIE-Net), as shown in~\cref{fig:model_framework22}, enhances low-light images using two inputs: an RGB image and its HSV-derived pixel-level features. The RGB input is downsampled and processed through five Self-Calibrated Pixel Attention (SCPA) blocks \cite{Zhang2022} for noise suppression and feature recalibration. These features are upsampled and fused with HSV features for color-aware enhancement \cite{Nathan2023}. A multi-stage encoder incorporating Residual Dense Attention (RDA) \cite{Uma2020} and NAFBlockSR \cite{Vasluianu2023} modules aggregates features and captures context. The decoder mirrors the encoder with upsampling and skip connections, using RDA blocks to refine textures. Auxiliary branches process the original input with attention mechanisms to preserve details and avoid over-smoothing. Outputs from the decoder, auxiliary branch, and a shallow pathway are fused with the original input to produce the final enhanced image. 

\noindent\textbf{Implementation:}  
LLIE-Net was trained on an NVIDIA Tesla P100 (16GB RAM) using TensorFlow Keras. It was trained on 4,281 patches (400$\times$400$\times$3) with random augmentations and validated on 755 patches. The Adam optimizer was used with a learning rate decaying from 0.001 to 0.00001 over 250 epochs.

%% file: team23_AVC2/main.tex
\subsection{AVC2}
\noindent\textbf{Description:}~We propose MobileIE, an efficient model for low-light image enhancement that balances parameters, speed, and performance. The model follows a simple design, utilizing basic operations in a streamlined topology (see \cref{AVC2}). During training, low-light images are processed through MBRConv 5$\times$5 and PReLU to extract shallow features. These features are then passed through two modules combining MBRConv 3$\times$3 and FST for further deep feature learning. An Attention module focuses on important regions, followed by MBRConv 3$\times$3 for fine processing to produce the final result. Inference uses reparameterized MBRConv layers.

\noindent\textbf{Implementation:}  
The model is implemented in PyTorch and tested on an RTX 3090 GPU. Optimization is done with the Adam optimizer and a cosine annealing learning rate schedule, starting at 0.001 and decaying every 50 epochs after a 10-epoch warm-up at 1$e$-6. Training lasts for 2,000 epochs, with the training data split into training and test sets (90/10 ratio) and progressively divided into patches of 500$\times$500, 1000$\times$1000, and 1500$\times$1500 in each stage.

%% file: team24_LR_LL/main.tex
\subsection{LR-LL}

\noindent\textbf{Description:}  
Our solution builds on LLNET~\cite{Fu_2022_CVPR} for low-light image enhancement in the YUV420 color space, as shown in \cref{fig:LR-LL-Model}. The approach consists of three key steps: 1) downscaling high-resolution low-light images and using a lightweight CNN to enhance them for fast processing, 2) applying guided upsampling to model the transformation between low-resolution input and output, and 3) using the estimated model to enhance high-resolution images, enabling real-time processing at full resolution. By performing most computations at lower resolution, the model achieves high-quality enhancement while reducing computational costs. Additionally, we introduce a dataset~\cite{aithal2025lenvizhighresolutionlowexposurenight} of low-light images with corresponding long-exposure references, captured in real-world conditions using smartphones.

\noindent\textbf{Implementation:}  
Implemented in TensorFlow and trained on a single NVIDIA A100 GPU using the NTIRE 2025 low-light enhancement dataset, the model uses the ADAM optimizer with a learning rate of 1$e$-4 for 1,000 epochs. The loss function is $\mathcal{L} = 2 \cdot \mathcal{L}_{1} + \mathcal{L}_{perc}$, where $\mathcal{L}_{1}$ is the mean absolute error and $\mathcal{L}_{perc}$ measures feature loss with a trained VGG-19 model. Designed for efficient smartphone deployment, the model can be converted to TFLite, utilizing the GPU delegate for processing. On the Qualcomm Adreno 735, it requires approximately 400MB of memory and processes a 2000×2992 image in around 100ms.

%% file: team25_X_L/main.tex
\subsection{X-L}
\noindent\textbf{Description:}~Inspired by the SYSU-FVL-T2 approach from the NTIRE-2024 low-light image enhancement challenge \cite{liu2024ntire}, we propose a method using ESDNet-L as the backbone. The backbone features an encoder-decoder network with three scales and skip connections, with features generated through bilinear interpolation.~At each scale, Semantic-Aligned Scale-Aware Modules are used to enhance scale variation handling, incorporating a pyramid context extraction module and cross-scale dynamic fusion for selective feature fusion.~Our modification adds permuted self-attention blocks after the SAM modules, improving the model’s ability to capture global dependencies and refine feature representations. For a detailed illustration, see \cref{fig:llie}.~The integration of self-attention with SAM improves the handling of scale variations, enhancing the model’s overall performance.

\noindent\textbf{Implementation:}  
We adopt a training strategy similar to SYSU-FVL-T2, using a single NVIDIA 4090 GPU.

%% file: team26_Team_IITRPR/main.tex
\subsection{Team\_IITRPR}
\noindent\textbf{Method description}:  
Our network, inspired by C2AIR \cite{kulkarni2024c2air}, consists of three modules, as illustrated in \cref{fig:TEAM_IITRPR}. The first component, the Degradation-Aware Query Modulated (DAQM) Block, adapts to varying lighting conditions by learning the degradation caused by underexposure. It modulates feature representations using illumination cues to emphasize dark regions needing enhancement. The second module, the Cross Collaborative Feed-Forward (CCF) Block, restores spatial details at multiple scales, ensuring both fine textures and large-scale structures are reconstructed. The third module, the Adaptive Gated Feature Fusion Block (AGFF), selectively integrates features across scales using a gating mechanism, suppressing noise and irrelevant content for naturally enhanced outputs.

\noindent\textbf{Implementation:}~The network is trained on the NTIRE 2025 challenge dataset using an NVIDIA GeForce RTX 1080 GPU with 8 GB of memory and a batch size of 1. The ADAM optimizer is used with a learning rate of 3 $\times$ 10$^{-4}$, $\beta_1$ = 0.5, and $\beta_2$ = 0.99. The model is trained for 110 epochs with random $512 \times 512$ patches and L1 loss.

%% file: team27_CV_SVNIT/main.tex
\subsection{CV-SVNIT (Excluded from Report)}

%% file: team28_JHC_INFO/main.tex
\subsection{JHC-INFO (Excluded from Ranking)}
\noindent\textbf{Description.}  
RetinexRWKV, as shown in \cref{fig:RetinexRWKV_backbone,fig:RetinexRWKV_unetblock}, is a lightweight model for low-light image enhancement, based on Retinexformer and RWKV TimeMix as shown in \cref{fig:RetinexRWKV_timemix}.~It efficiently integrates spatial and temporal information, supports up to 8K resolution with minimal computational load, and features linear attention with O(n) complexity for long-range dependencies.~Its dynamic state mechanism (RWKV v7) boosts adaptability and generalization across various visual enhancement tasks. More information can be found in~\cite{li2024retinexrwkv}.

\noindent\textbf{Implementation.}  
The model was trained using an AMD Radeon Pro W7900 GPU (48GB VRAM) with Triton acceleration in ROCm, showing strong hardware compatibility. However, pre-trained weights may not be reproducible on Nvidia systems, though training, forwarding, and testing remain feasible. With a batch size of 8, each epoch took 50 seconds, completing 100 epochs on the NTIRE 2025 dataset in 1.5 hours. To improve generalization and training speed, 256×256 random cropping was applied in batch training.

%% file: team01_NWPU_HVI/affiliation.tex
\subsection*{NWPU-HVI}
\noindent\textit{\textbf{Title: }}FusionNet: Multi-model Linear Fusion Framework for Low-light Image Enhancement \\
\noindent\textit{\textbf{Members: }} \\
Kangbiao Shi$^{1}$ (\href{mailto:18334840904@163.com}{18334840904@163.com})\\
Yixu Feng$^{1}$ (\href{mailto:yixu-nwpu@mail.nwpu.edu.cn}{yixu-nwpu@mail.nwpu.edu.cn})\\
Tao Hu$^{1}$ (\href{mailto:taohu@mail.nwpu.edu.cn}{taohu@mail.nwpu.edu.cn})\\
Yu Cao$^{2}$ (\href{mailto:caoyu@opt.ac.cn}{caoyu@opt.ac.cn})\\
Peng Wu$^{1}$ (\href{mailto:pengwu@nwpu.edu.cn}{pengwu@nwpu.edu.cn})\\
Yijin Liang$^{3}$ (\href{mailto:1719198363@qq.com}{1719198363@qq.com})\\
Yanning Zhang$^{1}$ (\href{mailto:ynzhang@nwpu.edu.cn}{ynzhang@nwpu.edu.cn})\\
Qingsen Yan$^{1}$ (\href{mailto:qingsenyan@nwpu.edu.cn}{qingsenyan@nwpu.edu.cn})\\
\noindent\textit{\textbf{Affiliations: }} \\ 
$^1$ School of Computer Science, Northwestern Polytechnical University, China \\
$^2$ Xi’an Institute of Optics and Precision Mechanics of CAS, China \\
$^3$ Shanghai Institute of Satellite Engineering, Shanghai 201109, China \\

%% file: team02_Imagine/affiliation.tex
\subsection*{Imagine}
\noindent\textit{\textbf{Title: }} SG-LLIE: Towards Scale-Aware Low-Light Enhancement via Structure-Guided Transformer Design \\
\noindent\textit{\textbf{Members: }} \\
Han Zhou~(\href{mailto:zhouh115@mcmaster.ca}{zhouh115@mcmaster.ca}) \\
Wei Dong~(\href{mailto:dongw22@mcmaster.ca}{dongw22@mcmaster.ca}) \\
Yan Min~(\href{mailto:miny13@mcmaster.ca}{miny13@mcmaster.ca}) \\
Mohab Kishawy~(\href{mailto:kishawym@mcmaster.ca}{kishawym@mcmaster.ca}) \\ 
Jun Chen~(\href{mailto:chenjun@mcmaster.ca}{chenjun@mcmaster.ca}) \\
\noindent\textit{\textbf{Affiliations: }} \\ 
McMaster University \\

%% file: team03_pengpeng_yu/affiliation.tex
\subsection*{pengpeng-yu}
\noindent\textit{\textbf{Title: }} Scale-Robust Low-Light Enhancement with Tricks
 \\
\noindent\textit{\textbf{Members: }} \\
Pengpeng Yu$^{1,2}$ (\href{mailto:yupp5@mail2.sysu.edu.cn}{yupp5@mail2.sysu.edu.cn})\\
\noindent\textit{\textbf{Affiliations: }} \\ 
$^1$ School of Electronics and Communication Engineering, Sun Yat-sen University, China \\
$^2$ Pengcheng Laboratory, China \\

%% file: team04_DAVIS_K/affiliation.tex
\subsection*{DAVIS-K}
\noindent\textit{\textbf{Title: }} Two-stage U-shaped Transformer for Low-light Image Enhancement \\
\noindent\textit{\textbf{Members: }} \\
Anjin Park$^{1}$ (\href{mailto:anjin.park@kopti.re.kr}{anjin.park@kopti.re.kr})\\ 
Seung-Soo Lee$^{2}$ (\href{mailto:sslee1@doowoninc.com}{sslee1@doowoninc.com})\\
Young-Joon Park$^{2}$ (\href{mailto:yjpark@doowoninc.com}{yjpark@doowoninc.com})\\
\noindent\textit{\textbf{Affiliations: }} \\ 
$^1$ Intelligent Technology Research Center, Korea Photonics Technology Institute, Korea \\
$^2$ Doowon Electronics and Telecom Co., LTD., Korea \\

%% file: team05_SoloMan/affiliation.tex
\subsection*{SoloMan}
\noindent\textit{\textbf{Title: }} ESDNet-Twins: Twin network using complementary Ensemble strategy \\
\noindent\textit{\textbf{Members: }} \\
Zixiao Hu~(\href{mailto:1812535642@qq.com}{1812535642@qq.com}),
Junyv Liu~(\href{mailto:1062973385@qq.com}{1062973385@qq.com}),
Huilin Zhang~(\href{mailto:huilin0914@163.com}{huilin0914@163.com}),
Jun Zhang~(\href{mailto:202411081539@std.uestc.edu.cn}{202411081539@std.uestc.edu.cn}),\\
\noindent\textit{\textbf{Affiliations: }} \\ 
University of Electronic Science and Technology of China, China \\

%% file: team06_Smartdsp/affiliation.tex

%% file: team07_Smart210/affiliation.tex
\subsection*{Smart210}
\noindent\textit{\textbf{Title: }}ESDNet with SimPF block for low light image enhancement \\
\noindent\textit{\textbf{Members: }} \\
Fei Wan$^{1,2}$ (\href{mailto:mywsi566131@outlook.com}{mywsi566131@outlook.com}), Bingxin Xu$^{1,2}$, Hongzhe Liu$^{1,2}$, Cheng Xu$^{1,2}$, Weiguo Pan$^{1,2}$, Songyin Dai$^{1,2}$\\
\noindent\textit{\textbf{Affiliations: }} \\ 
$^1$ Beijing Key Laboratory of Information Service Engineering, Beijing Union University, Beijing, China \\
$^2$ College of Robotics, Beijing Union University, Beijing, China \\

%% file: team08_WHU_MVP/affiliation.tex
\subsection*{WHU-MVP}
\noindent\textit{\textbf{Title: }} Hierarchical Coarse-to-Fine Network for Low-Light Image Enhancement \\
\noindent\textit{\textbf{Members: }} \\
Xunpeng Yi~(\href{mailto:xpyi2008@163.com}{xpyi2008@163.com}), \\
Qinglong Yan~(\href{mailto:qinglong_yan@whu.edu.cn}{qinglong\_yan@whu.edu.cn}), \\ 
Yibing Zhang~(\href{zhangyibing@whu.edu.cn}{zhangyibing@whu.edu.cn}), \\ 
Jiayi Ma~(\href{jyma2010@gmail.com}{jyma2010@gmail.com}), \\
\noindent\textit{\textbf{Affiliations: }} \\ 
Electronic Information School, Wuhan University, China \\

%% file: team09_BUPTMM/affiliation.tex

%% file: team10_NJUPTIPR/affiliation.tex
\subsection*{NJUPT-IPR}
\noindent\textit{\textbf{Title: }}Enhanced IIAG-CoFlow: Inter-and Intra-channel Attention Transformer and Complete Flow for Low-light Image Enhancement \\
\noindent\textit{\textbf{Members: }} \\
Changhui Hu~(\href{mailto:hchnjupt@126.com}{hchnjupt@126.com})\\
Kerui Hu~(\href{mailto:hukerui7@gmail.com}{hukerui7@gmail.com})\\
Donghang Jing~(\href{mailto:1023051419@njupt.edu.cn}{1023051419@njupt.edu.cn})\\
Tiesheng Chen~(\href{mailto:1223056018@njupt.edu.cn}{1223056018@njupt.edu.cn})\\
\noindent\textit{\textbf{Affiliations: }} \\ 
College of Automation and College of Artificial Intelligence, Nanjing University of Posts and Telecommunications, Nanjing 210023, China \\

%% file: team11_SYSU_FVL_T2/affiliation.tex
\subsection*{SYSU-FVL-T2}
\noindent\textit{\textbf{Title:}} Improved Scale-Robust Low-light Ultra-High-Definition Image Enhancement \\
\noindent\textit{\textbf{Members: }} \\
Zhi Jin$^{1,2}$ (\href{mailto:xxxxxxxxxxxxx}{jinzh26@mail.sysu.edu.cn}),
Hongjun Wu$^1$, 
Biao Huang$^1$, 
Haitao Ling$^1$,
Jiahao Wu$^1$,
Dandan Zhan$^1$\\
\noindent\textit{\textbf{Affiliations: }} \\ 
$^1$School of Intelligent Systems Engineering, Shenzhen Campus of Sun Yat-sen University, Shenzhen, Guangdong 518107, China \\
$^2$Guangdong Provincial Key Laboratory of Fire Science and Intelligent Emergency Technology, Guangzhou 510006, China\\

%% file: team12_KLETech_CEVI/affiliation.tex
\subsection*{ESDNet+: Enhanced Scale-Robust Network for Low-Light Enhancement}
\noindent\textit{\textbf{Title: }} ESDNet+: Enhanced Scale-Robust Network for Low-Light Enhancement \\
\noindent\textit{\textbf{Members: }} \\
G Gyaneshwar Rao$^{2,3}$ (\href{mailto:01fe21bec172@kletech.ac.in}{01fe21bec172@kletech.ac.in}), Vijayalaxmi Ashok Aralikatti$^{1,3}$, Nikhil Akalwadi$^{1,3}$, Ramesh Ashok Tabib$^{2,3}$, Uma Mudenagudi$^{2,3}$\\
\noindent\textit{\textbf{Affiliations: }} \\ 
$^{1}$ School of Computer Science and Engineering, KLE Technological University\\
$^{2}$ School of Electronics and Communication Engineering, KLE Technological University\\
$^{3}$ Center of Excellence in Visual Intelligence (CEVI), KLE Technological University\\

%% file: team13_Ensemble_KNights/affiliation.tex
\subsection*{Ensemble-KNights}
\noindent\textit{\textbf{Title: }} Transformer or Mamba: Can an Ensemble of Both Shine Brighter in Low-Light Image Enhancement? \\
\noindent\textit{\textbf{Members: }} \\
Ruirui Lin$^{1}$ (\href{mailto:r.lin@bristol.ac.uk}{r.lin@bristol.ac.uk})\\
Guoxi Huang$^{1}$ (\href{mailto:guoxi.huang@bristol.ac.uk}{guoxi.huang@bristol.ac.uk})\\
Nantheera Anantrasirichai$^{1}$ (\href{mailto:N.Anantrasirichai@bristol.ac.uk}{N.Anantrasirichai@bristol.ac.uk})\\
Qirui Yang$^{2}$ (\href{mailto:yangqirui@tju.edu.cn}{yangqirui@tju.edu.cn})\\

\noindent\textit{\textbf{Affiliations:}} \\ 
$^1$ Visual Information Laboratory, University of Bristol, Bristol, UK \\
$^2$ School of Electrical and Information Engineering, Tianjin University, China \\

%% file: team14_MRT_LLIE/affiliation.tex
\subsection*{MRT-LLIE}
\noindent\textit{\textbf{Title: }} Multi-stage Residual Transformer for Low-light Image Enhancement \\
\noindent\textit{\textbf{Members: }} \\
Alexandru Brateanu$^{1}$ \\(\href{mailto:alexandru.brateanu@student.manchester.ac.uk}{alexandru.brateanu@student.manchester.ac.uk})\\
Ciprian Orhei$^{2}$ (\href{mailto:ciprian.orhei@upt.ro}{ciprian.orhei@upt.ro})\\
Cosmin Ancuti$^{2}$ (\href{mailto:cosmin.ancuti@upt.ro}{cosmin.ancuti@upt.ro})\\
\noindent\textit{\textbf{Affiliations: }} \\ 
$^1$ University of Manchester, Manchester, United Kingdom \\
$^2$ Polytechnic University Timisoara, Timisoara, Romania\\

%% file: team15_SynLLIE/affiliation.tex

%% file: team16_CidautAI/affiliation.tex
\subsection*{Cidaut AI}
\noindent\textit{\textbf{Title: }} DarkIR and FLOL  \\
\noindent\textit{\textbf{Members: }} \\
Daniel Feijoo$^{1}$,
Juan C. Benito$^{1}$,
Álvaro García$^{1}$,
Marcos V. Conde$^{1,2}$(\href{mailto:marcos.conde}{marcos.conde@uni-wuerzburg.de})\\
\noindent\textit{\textbf{Affiliations: }} \\ 
$^1$ Cidaut AI, Spain \\
$^2$ Computer Vision Lab, University of W\"urzburg, Germany  \\

%% file: team17_Huabujianye/affiliation.tex
\subsection*{Huabujianye}
\noindent\textit{\textbf{Title: }} D-RetinexMix \\
\noindent\textit{\textbf{Members: }} \\
Yang Qin~(\href{mailto:1870602063@qq.com}{1870602063@qq.com})\\
\noindent\textit{\textbf{Affiliations:}} \\ 
Nanjing University of Information Science and Technology \\

%% file: team18_no_way_no_lay/affiliation.tex
\subsection*{no way no lay}
\noindent\textit{\textbf{Title: }} retimixformer: Retinex-guided Multi-branch Transformer for Low-Light Image Enhancement \\
\noindent\textit{\textbf{Members: }} \\
Yufeng Yang$^{1}$~(\href{mailto:dericky286@gmail.com}{dericky286@gmail.com}), 
Jiancheng Huang$^{2}$, 
Donghao Zhou$^{3}$, 
Shifeng Chen$^{1}$ \\
\noindent\textit{\textbf{Affiliations: }} \\
$^{1}$ Southern University of Science and Technology, Shenzhen, China \\
$^{2}$ Shenzhen Institute of Advanced Technology, Chinese Academy of Sciences, China \\
$^{3}$ The Chinese University of Hong Kong, Hong Kong \\

%% file: team19_Lux_Themps/affiliation.tex
\subsection*{Lux Themps}
\noindent\textit{\textbf{Title: }} SADe-ViT: Semantic-Aware Depthwise Vision Transformer for Low Light Image Enhancement \\
\noindent\textit{\textbf{Members: }} \\
Raul Balmez$^{1}$ (\href{mailto:raul.balmez@student.manchester.ac.uk}{raul.balmez@student.manchester.ac.uk}),
Cosmin Ancuti$^{2}$ (\href{mailto:cosmin.ancuti@upt.ro}{cosmin.ancuti@upt.ro}), \\
Ciprian Orhei$^{2}$ (\href{mailto:ciprian.orhei@upt.ro}{ciprian.orhei@upt.ro})\\
\noindent\textit{\textbf{Affiliations: }} \\ 
$^1$ University of Manchester, Manchester, United Kingdom \\
$^2$ Polytechnic University Timisoara, Timisoara, Romania \\

%% file: team20_PSU_TEAM/affiliation.tex
\subsection*{PSU\_team}
\noindent\textit{\textbf{Title: }} OptimalDiff: High-Fidelity Image Enhancement Using Schrödinger Bridge Diffusion and Multi-Scale Adversarial Refinement\\

\noindent\textit{\textbf{Members: }} \\
Anas M. Ali~(\href{mailto:aaboessa@psu.edu.sa}{aaboessa@psu.edu.sa}), 
Bilel Benjdira~(\href{mailto:bbenjdira@psu.edu.sa}{bbenjdira@psu.edu.sa}), 
Wadii Boulila~(\href{mailto:wboulila@psu.edu.sa}{wboulila@psu.edu.sa})\\

\noindent\textit{\textbf{Affiliations: }} \\ 
Robotics and Internet-of-Things Laboratory, Prince Sultan University, Riyadh, Saudi Arabia \\

%% file: team21_hfut_lvgroup/affiliation.tex
\subsection*{hfut-lvgroup}
\noindent\textit{\textbf{Title: }} Enhanced RetinexFormer \\
\noindent\textit{\textbf{Members: }} \\
Tianyi Mao$^1$ (\href{mailto:mty114514@gmail.com}{mty114514@gmail.com})\\
Huan Zheng$^2$ (\href{mailto:yc37928@umac.mo}{yc37928@umac.mo})\\
Yanyan Wei$^1$ (\href{mailto:weiyy@hfut.edu.cn}{weiyy@hfut.edu.cn})\\
Shengeng Tang$^1$ (\href{mailto:tangsg@hfut.edu.cn}{tangsg@hfut.edu.cn})\\
Dan Guo$^1$ (\href{mailto:guodan@hfut.edu.cn}{guodan@hfut.edu.cn})\\
Zhao Zhang$^1$ (\href{mailto:cszzhang@gmail.com}{cszzhang@gmail.com})\\
\noindent\textit{\textbf{Affiliations: }} \\ 
$^1$ School of Computer and Information, Hefei University of Technology \\
$^2$ University of Macau \\

%% file: team22_ImageLab/affiliation.tex
\subsection*{ImageLab}
\noindent\textit{\textbf{Title: }} Lightweight Self-Calibrated Pixel-Attentive
Network for Low-Light Image Enhancement \\
\noindent\textit{\textbf{Members: }} \\
Sabari Nathan$^1$ (\href{mailto:sabarinathan@gmail.com}{sabarinathantce@gmail.com})\\
K Uma$^2$ (\href{uma.k@sasi.ac.in}{<uma.k@sasi.ac.in})\\
A Sasithradevi$^3$(\href{sasithradevi.a@vit.ac.in}{sasithradevi.a@vit.ac.in})\\
B Sathya Bama$^4$ (\href{mailto:sbece@tce.edu}{sbece@tce.edu})\\
S. Mohamed Mansoor Roomi$^4$(\href{smmroomi@tce.edu}{smmroomi@tce.edu})\\
\noindent\textit{\textbf{Affiliations: }} \\ 
$^1$ Couger In ,Japan \\
$^2$ Sasi Institute of Technology \& Engineering, India \\
$^3$ Vellore Institute of Technology, India\\
$^4$ Thiagarajar college of engineering, India \\

%% file: team23_AVC2/affiliation.tex
\subsection*{AVC2}
\noindent\textit{\textbf{Title:}}~MobileIE: Towards Efficient Real-Time Image Enhancement for Mobile Devices. \\
\noindent\textit{\textbf{Members: }} \\
Ao Li$^1$ (\href{mailto:aoli@std.uestc.edu.cn}{aoli@std.uestc.edu.cn})\\
Xiangtao Zhang$^1$ (\href{mailto:xiangtaozhang@std.uestc.edu.cn}{xiangtaozhang@std.uestc.edu.cn})\\
Zhe Liu$^1$ (\href{mailto:liuzhe@std.uestc.edu.cn}{liuzhe@std.uestc.edu.cn})\\
Yijie Tang$^2$ (\href{mailto:tangyijie@std.cqu.edu.cn}{tangyijie@std.cqu.edu.cn})\\
Jialong Tang$^3$ (\href{mailto:Tangjialong20@163.com}{Tangjialong20@163.com})\\
\noindent\textit{\textbf{Affiliations: }} \\ 
$^1$~University of Electronic Science and Technology of
China, China \\
$^2$~Chongqing University, China \\
$^3$~Zhejiang University, China \\

%% file: team24_LR_LL/affiliation.tex
\subsection*{LR-LL}
\noindent\textit{\textbf{Title: }} An Efficient AI Solution for Low-Light Image Enhancement on Mobile Devices \\
\noindent\textit{\textbf{Members: }} \\
Zhicheng Fu~(\href{mailto:zcfu@lenovo.com}{zcfu@lenovo.com})\\
Gong Chen~(\href{mailto:gochen24@motorola.com}{gochen24@motorola.com})\\ 
Joe Nasti~(\href{mailto:joenasti@motorola.com}{joenasti@motorola.com})\\
John Nicholson~(\href{mailto:jnichol@lenovo.com}{jnichol@lenovo.com})\\
\noindent\textit{\textbf{Affiliations:}} \\
Lenovo Research\\

%% file: team25_X_L/affiliation.tex
\subsection*{X-L}
\noindent\textit{\textbf{Title: }} Enhanced ESDNet\\
\noindent\textit{\textbf{Members: }} \\
Zeyu Xiao$^1$ (\href{mailto:zeyuxiao1997@163.com}{zeyuxiao1997@163.com})\\
Zhuoyuan Li$^2$  (\href{mailto:zhuoyuanli@mail.ustc.edu.cn}{zhuoyuanli@mail.ustc.edu.cn})\\
\noindent\textit{\textbf{Affiliations: }} \\ 
$^1$ National University of Singapore \\
$^2$ University of Science and Technology of China \\

%% file: team26_Team_IITRPR/affiliation.tex
\subsection*{Team\_IITRPR}
\noindent\textit{\textbf{Title: }}IllumiNet: A Lightweight Approach for Lowlight Image Enhancement\\
\noindent\textit{\textbf{Members: }} \\
Ashutosh Kulkarni$^{1}$ (\href{mailto:}{ashutosh.20eez0008@iitrpr.ac.in})\\
Prashant W. Patil$^{2}$ (\href{mailto:}{pwpatil@iitg.ac.in})\\
Santosh Kumar Vipparthi$^{1}$ (\href{mailto:}{skvipparthi@iitrpr.ac.in})\\
Subrahmanyam Murala$^{3}$ (\href{mailto:}{muralas@tcd.ie})\\
\noindent\textit{\textbf{Affiliations: }} \\ 
$^1$CVPR Lab, Indian Institute of Technology Ropar, India \\
$^2$MFSDSAI, Indian Institute of Technology Guwahati, India\\
$^3$CVPR Lab, School of Computer Science and Statistics, Trinity College Dublin, Dublin, Ireland \\

%% file: team27_CV_SVNIT/affiliation.tex

%% file: team28_JHC_INFO/affiliation.tex
\subsection*{JHC-Info}
\noindent\textit{\textbf{Title: }} RetinexRWKV: A Streamlined Linear Attention Mechanism to Substitute Self-Attention \\
\noindent\textit{\textbf{Members: }} \\
Duan Liu$^{1}$ (\href{mailto:liuduanchn@jhc.edu.cn}{liuduanchn@jhc.edu.cn})\\
Weile Li$^{1}$ (\href{mailto:alic2591709191@gmail.com}{alic2591709191@gmail.com})\\
Hangyuan Lu$^{1}$ (\href{mailto:lhyhziee@163.com}{lhyhziee@163.com})\\
Rixian Liu$^{1}$  (\href{mailto:liurixian@163.com}{liurixian@163.com})\\
Tengfeng Wang$^{2}$ (\href{mailto:wangtf@cqupt.edu.cn}{wangtf@cqupt.edu.cn})\\
Jinxing Liang$^{3}$ (\href{mailto:jxliang@wtu.edu.cn}{jxliang@wtu.edu.cn})\\
Chenxin Yu$^{3}$ (\href{mailto:3505756463@qq.com}{3505756463@qq.com})\\
\noindent\textit{\textbf{Affiliations: }} \\ 
$^1$ Infomation Engineering College, Jinhua University of Vocational Technology \\
$^2$ School of Software Engineering, Chongqing University of Posts and Telecommunications\\
$^3$ School of Computer Science and Artificial Intelligence,Wuhan Textile University\\

%% file: main.bbl
\begin{thebibliography}{10}\itemsep=-1pt

\bibitem{aithal2025lenvizhighresolutionlowexposurenight}
Manjushree Aithal, Rosaura~G. VidalMata, Manikandtan Kartha, Gong Chen, Eashan
  Adhikarla, Lucas~N. Kirsten, Zhicheng Fu, Nikhil~A. Madhusudhana, and Joe
  Nasti.
\newblock Lenviz: A high-resolution low-exposure night vision benchmark
  dataset, 2025.

\bibitem{depthlux_isetc}
Raul Balmez and Alexandru Brateanu.
\newblock Depthlux: Depthwise separable convolution transformer.
\newblock In {\em 2024 International Symposium on Electronics and
  Telecommunications (ISETC)}, pages 1--4, 2024.

\bibitem{depthlux_mdpi}
Raul Balmez, Alexandru Brateanu, Ciprian Orhei, Codruta~O. Ancuti, and Cosmin
  Ancuti.
\newblock Depthlux: Employing depthwise separable convolutions for low-light
  image enhancement.
\newblock {\em Sensors}, 25(5), 2025.

\bibitem{benito2025flol}
Juan~C Benito, Daniel Feijoo, Alvaro Garcia, and Marcos~V Conde.
\newblock Flol: Fast baselines for real-world low-light enhancement.
\newblock {\em arXiv preprint arXiv:2501.09718}, 2025.

\bibitem{lytnet2024}
Alexandru Brateanu, Raul Balmez, Adrian Avram, Ciprian Orhei, and Ancuti
  Cosmin.
\newblock Lyt-net: Lightweight yuv transformer-based network for low-light
  image enhancement.
\newblock {\em arXiv preprint arXiv:2401.15204}, 2024.

\bibitem{retinexformer}
Yuanhao Cai, Hao Bian, Jing Lin, Haoqian Wang, Radu Timofte, and Yulun Zhang.
\newblock Retinexformer: One-stage retinex-based transformer for low-light
  image enhancement.
\newblock In {\em Proceedings of the IEEE/CVF International Conference on
  Computer Vision}, pages 12504--12513, 2023.

\bibitem{chen2022simple}
Liangyu Chen, Xiaojie Chu, Xiangyu Zhang, and Jian Sun.
\newblock Simple baselines for image restoration.
\newblock In {\em European conference on computer vision}, pages 17--33.
  Springer, 2022.

\bibitem{MobilNetV3}
Liang-Chieh Chen, George Papandreou, Florian Schroff, and Hartwig Adam.
\newblock Rethinking atrous convolution for semantic image segmentation.
\newblock {\em arXiv preprint arXiv:1706.05587}, 2017.

\bibitem{ntire2025srx4}
Zheng Chen, Kai Liu, Jue Gong, Jingkai Wang, Lei Sun, Zongwei Wu, Radu Timofte,
  Yulun Zhang, et~al.
\newblock {NTIRE} 2025 challenge on image super-resolution (×4): Methods and
  results.
\newblock In {\em Proceedings of the IEEE/CVF Conference on Computer Vision and
  Pattern Recognition (CVPR) Workshops}, 2025.

\bibitem{ntire2025face}
Zheng Chen, Jingkai Wang, Kai Liu, Jue Gong, Lei Sun, Zongwei Wu, Radu Timofte,
  Yulun Zhang, et~al.
\newblock {NTIRE} 2025 challenge on real-world face restoration: Methods and
  results.
\newblock In {\em Proceedings of the IEEE/CVF Conference on Computer Vision and
  Pattern Recognition (CVPR) Workshops}, 2025.

\bibitem{ntire2025raw}
Marcos Conde, Radu Timofte, et~al.
\newblock {NTIRE} 2025 challenge on raw image restoration and super-resolution.
\newblock In {\em Proceedings of the IEEE/CVF Conference on Computer Vision and
  Pattern Recognition (CVPR) Workshops}, 2025.

\bibitem{ntire2025rawrgb}
Marcos Conde, Radu Timofte, et~al.
\newblock Raw image reconstruction from {RGB} on smartphones. {NTIRE} 2025
  challenge report.
\newblock In {\em Proceedings of the IEEE/CVF Conference on Computer Vision and
  Pattern Recognition (CVPR) Workshops}, 2025.

\bibitem{Vasluianu2023}
M.V. Conde, F. Vasluianu, S. Nathan, and R. Timofte.
\newblock Real-time under-display cameras image restoration and hdr on mobile
  devices.
\newblock In {\em Computer Vision – ECCV 2022 Workshops}, volume 13802 of
  {\em Lecture Notes in Computer Science}. Springer, Cham, 2023.

\bibitem{ntire2025night}
Egor Ershov, Sergey Korchagin, Alexei Khalin, Artyom Panshin, Arseniy Terekhin,
  Ekaterina Zaychenkova, Georgiy Lobarev, Vsevolod Plokhotnyuk, Denis Abramov,
  Elisey Zhdanov, Sofia Dorogova, Yasin Mamedov, Nikola Banic, Georgii
  Perevozchikov, Radu Timofte, et~al.
\newblock {NTIRE} 2025 challenge on night photography rendering.
\newblock In {\em Proceedings of the IEEE/CVF Conference on Computer Vision and
  Pattern Recognition (CVPR) Workshops}, 2025.

\bibitem{feijoo2024darkir}
Daniel Feijoo, Juan~C Benito, Alvaro Garcia, and Marcos~V Conde.
\newblock Darkir: Robust low-light image restoration.
\newblock {\em arXiv preprint arXiv:2412.13443}, 2024.

\bibitem{feng2022mipi}
Ruicheng Feng, Chongyi Li, Shangchen Zhou, Wenxiu Sun, Qingpeng Zhu, Jun Jiang,
  Qingyu Yang, Chen~Change Loy, Jinwei Gu, Yurui Zhu, et~al.
\newblock Mipi 2022 challenge on under-display camera image restoration:
  Methods and results.
\newblock In {\em European Conference on Computer Vision}, pages 60--77.
  Springer, 2022.

\bibitem{ntire2025cross}
Yuqian Fu, Xingyu Qiu, Bin Ren~Yanwei Fu, Radu Timofte, Nicu Sebe, Ming-Hsuan
  Yang, Luc Van~Gool, et~al.
\newblock {NTIRE} 2025 challenge on cross-domain few-shot object detection:
  Methods and results.
\newblock In {\em Proceedings of the IEEE/CVF Conference on Computer Vision and
  Pattern Recognition (CVPR) Workshops}, 2025.

\bibitem{Fu_2022_CVPR}
Zhicheng Fu, Miao Song, Chao Ma, Joseph Nasti, Vivek Tyagi, Grant Lloyd, and
  Wei Tang.
\newblock An efficient hybrid model for low-light image enhancement in mobile
  devices.
\newblock In {\em Proceedings of the IEEE/CVF Conference on Computer Vision and
  Pattern Recognition (CVPR) Workshops}, pages 3057--3066, June 2022.

\bibitem{ntire2025text}
Shuhao Han, Haotian Fan, Fangyuan Kong, Wenjie Liao, Chunle Guo, Chongyi Li,
  Radu Timofte, et~al.
\newblock {NTIRE} 2025 challenge on text to image generation model quality
  assessment.
\newblock In {\em Proceedings of the IEEE/CVF Conference on Computer Vision and
  Pattern Recognition (CVPR) Workshops}, 2025.

\bibitem{10845035}
Changhui Hu, Tiesheng Chen, Donghang Jing, Kerui Hu, Yanyong Guo, Xiaoyuan
  Jing, and Pan Liu.
\newblock Iiag-coflow: Inter-and intra-channel attention transformer and
  complete flow for low-light image enhancement with application to night
  traffic monitoring images.
\newblock {\em IEEE Transactions on Intelligent Transportation Systems}, pages
  1--19, 2025.

\bibitem{huang2025bayesian}
Guoxi Huang, Nantheera Anantrasirichai, Fei Ye, Zipeng Qi, RuiRui Lin, Qirui
  Yang, and David Bull.
\newblock Bayesian neural networks for one-to-many mapping in image
  enhancement.
\newblock {\em arXiv preprint arXiv:2501.14265}, 2025.

\bibitem{ntire2025vqe}
Varun Jain, Zongwei Wu, Quan Zou, Louis Florentin, Henrik Turbell, Sandeep
  Siddhartha, Radu Timofte, et~al.
\newblock {NTIRE} 2025 challenge on video quality enhancement for video
  conferencing: Datasets, methods and results.
\newblock In {\em Proceedings of the IEEE/CVF Conference on Computer Vision and
  Pattern Recognition (CVPR) Workshops}, 2025.

\bibitem{jiang2023low}
Hai Jiang, Ao Luo, Haoqiang Fan, Songchen Han, and Shuaicheng Liu.
\newblock Low-light image enhancement with wavelet-based diffusion models.
\newblock {\em ACM Transactions on Graphics (TOG)}, 42(6):1--14, 2023.

\bibitem{adam}
Diederik~P. Kingma and Jimmy~Lei Ba.
\newblock Adam: A method for stochastic optimization.
\newblock In {\em ICLR}, 2015.

\bibitem{kulkarni2024c2air}
Ashutosh Kulkarni, Shruti~S Phutke, Santosh~Kumar Vipparthi, and Subrahmanyam
  Murala.
\newblock C2air: consolidated compact aerial image haze removal.
\newblock In {\em Proceedings of the IEEE/CVF Winter Conference on Applications
  of Computer Vision}, pages 749--758, 2024.

\bibitem{chan}
Wei-Sheng Lai, Jia-Bin Huang, Narendra Ahuja, and Ming-Hsuan Yang.
\newblock Fast and accurate image super-resolution with deep laplacian pyramid
  networks.
\newblock {\em IEEE transactions on pattern analysis and machine intelligence},
  41(11):2599--2613, 2018.

\bibitem{vggloss}
Christian Ledig, Lucas Theis, Ferenc Husz{'a}r, Jose Caballero, Andrew
  Cunningham, Alejandro Acosta, Andrew Aitken, Alykhan Tejani, Johannes Totz,
  Zehan Wang, et~al.
\newblock Photo-realistic single image super-resolution using a generative
  adversarial network.
\newblock In {\em Proceedings of the IEEE conference on computer vision and
  pattern recognition}, pages 4681--4690, 2017.

\bibitem{ntire2025ebhdr}
Sangmin Lee, Eunpil Park, Angel Canelo, Hyunhee Park, Youngjo Kim, Hyungju
  Chun, Xin Jin, Chongyi Li, Chun-Le Guo, Radu Timofte, et~al.
\newblock {NTIRE} 2025 challenge on efficient burst hdr and restoration:
  Datasets, methods, and results.
\newblock In {\em Proceedings of the IEEE/CVF Conference on Computer Vision and
  Pattern Recognition (CVPR) Workshops}, 2025.

\bibitem{li2024retinexrwkv}
Weile Li, Duan Liu, Hangyuan Lu, Rixian Liu, Tengfeng Wang, Jinxing Liang, and
  Chenxin Yu.
\newblock Retinexrwkv: A streamlined linear attention mechanism to substitute
  self-attention, 2025.
\newblock Submitted to arXiv, Submission ID: 6345948, April 2025.

\bibitem{ntire2025day}
Xin Li, Yeying Jin, Xin Jin, Zongwei Wu, Bingchen Li, Yufei Wang, Wenhan Yang,
  Yu Li, Zhibo Chen, Bihan Wen, Robby Tan, Radu Timofte, et~al.
\newblock {NTIRE} 2025 challenge on day and night raindrop removal for
  dual-focused images: Methods and results.
\newblock In {\em Proceedings of the IEEE/CVF Conference on Computer Vision and
  Pattern Recognition (CVPR) Workshops}, 2025.

\bibitem{ntire2025shortugc_data}
Xin Li, Xijun Wang, Bingchen Li, Kun Yuan, Yizhen Shao, Suhang Yao, Ming Sun,
  Chao Zhou, Radu Timofte, and Zhibo Chen.
\newblock {NTIRE} 2025 challenge on short-form ugc video quality assessment and
  enhancement: Kwaisr dataset and study.
\newblock In {\em Proceedings of the IEEE/CVF Conference on Computer Vision and
  Pattern Recognition (CVPR) Workshops}, 2025.

\bibitem{ntire2025shortugc}
Xin Li, Kun Yuan, Bingchen Li, Fengbin Guan, Yizhen Shao, Zihao Yu, Xijun Wang,
  Yiting Lu, Wei Luo, Suhang Yao, Ming Sun, Chao Zhou, Zhibo Chen, Radu
  Timofte, et~al.
\newblock {NTIRE} 2025 challenge on short-form ugc video quality assessment and
  enhancement: Methods and results.
\newblock In {\em Proceedings of the IEEE/CVF Conference on Computer Vision and
  Pattern Recognition (CVPR) Workshops}, 2025.

\bibitem{ntire2025raim}
Jie Liang, Radu Timofte, Qiaosi Yi, Zhengqiang Zhang, Shuaizheng Liu, Lingchen
  Sun, Rongyuan Wu, Xindong Zhang, Hui Zeng, Lei Zhang, et~al.
\newblock {NTIRE} 2025 the 2nd restore any image model {(RAIM)} in the wild
  challenge.
\newblock In {\em Proceedings of the IEEE/CVF Conference on Computer Vision and
  Pattern Recognition (CVPR) Workshops}, 2025.

\bibitem{ntire2025xgc}
Xiaohong Liu, Xiongkuo Min, Qiang Hu, Xiaoyun Zhang, Jie Guo, et~al.
\newblock {NTIRE} 2025 {XGC} quality assessment challenge: Methods and results.
\newblock In {\em Proceedings of the IEEE/CVF Conference on Computer Vision and
  Pattern Recognition (CVPR) Workshops}, 2025.

\bibitem{liu2024ntire}
Xiaoning Liu, Zongwei Wu, Ao Li, Florin-Alexandru Vasluianu, Yulun Zhang,
  Shuhang Gu, Le Zhang, Ce Zhu, Radu Timofte, Zhi Jin, et~al.
\newblock Ntire 2024 challenge on low light image enhancement: Methods and
  results.
\newblock In {\em Proceedings of the IEEE/CVF Conference on Computer Vision and
  Pattern Recognition}, pages 6571--6594, 2024.

\bibitem{liu2022convnet}
Zhuang Liu, Hanzi Mao, Chao-Yuan Wu, Christoph Feichtenhofer, Trevor Darrell,
  and Saining Xie.
\newblock A convnet for the 2020s.
\newblock In {\em Proceedings of the IEEE/CVF conference on computer vision and
  pattern recognition}, pages 11976--11986, 2022.

\bibitem{sgdr}
Ilya Loshchilov and Frank Hutter.
\newblock Sgdr: Stochastic gradient descent with warm restarts.
\newblock In {\em ICLR}, 2017.

\bibitem{Uma2020}
D. Nathan, K. Uma, D.~S. Vinothini, B.~S. Bama, and S.~M.~M. Roomi.
\newblock Lightweight residual dense attention net for spectral reconstruction
  from rgb images.
\newblock In {\em arXiv preprint arXiv:2004.06930}, 2020.
\newblock Available at: \url{https://arxiv.org/abs/2004.06930}.

\bibitem{Nathan2023}
S. Nathan and P. Kansal.
\newblock End-to-end depth-guided relighting using lightweight deep
  learning-based method.
\newblock In {\em Proceedings of the Journal of Imaging}, volume~9, page 175,
  2023.

\bibitem{ntire2025esr}
Bin Ren, Hang Guo, Lei Sun, Zongwei Wu, Radu Timofte, Yawei Li, et~al.
\newblock The tenth {NTIRE} 2025 efficient super-resolution challenge report.
\newblock In {\em Proceedings of the IEEE/CVF Conference on Computer Vision and
  Pattern Recognition (CVPR) Workshops}, 2025.

\bibitem{ntire2025ugc}
Nickolay Safonov, Alexey Bryntsev, Andrey Moskalenko, Dmitry Kulikov, Dmitriy
  Vatolin, Radu Timofte, et~al.
\newblock {NTIRE} 2025 challenge on {UGC} video enhancement: Methods and
  results.
\newblock In {\em Proceedings of the IEEE/CVF Conference on Computer Vision and
  Pattern Recognition (CVPR) Workshops}, 2025.

\bibitem{vgg19}
Karen Simonyan and Andrew Zisserman.
\newblock Very deep convolutional networks for large-scale image recognition.
\newblock {\em arXiv preprint arXiv:1409.1556}, 2014.

\bibitem{ntire2025event}
Lei Sun, Andrea Alfarano, Peiqi Duan, Shaolin Su, Kaiwei Wang, Boxin Shi, Radu
  Timofte, Danda~Pani Paudel, Luc Van~Gool, et~al.
\newblock {NTIRE} 2025 challenge on event-based image deblurring: Methods and
  results.
\newblock In {\em Proceedings of the IEEE/CVF Conference on Computer Vision and
  Pattern Recognition (CVPR) Workshops}, 2025.

\bibitem{ntire2025denoising}
Lei Sun, Hang Guo, Bin Ren, Luc Van~Gool, Radu Timofte, Yawei Li, et~al.
\newblock The tenth ntire 2025 image denoising challenge report.
\newblock In {\em Proceedings of the IEEE/CVF Conference on Computer Vision and
  Pattern Recognition (CVPR) Workshops}, 2025.

\bibitem{ntire2025shadow}
Florin-Alexandru Vasluianu, Tim Seizinger, Zhuyun Zhou, Cailian Chen, Zongwei
  Wu, Radu Timofte, et~al.
\newblock {NTIRE} 2025 image shadow removal challenge report.
\newblock In {\em Proceedings of the IEEE/CVF Conference on Computer Vision and
  Pattern Recognition (CVPR) Workshops}, 2025.

\bibitem{ntire2025ambient}
Florin-Alexandru Vasluianu, Tim Seizinger, Zhuyun Zhou, Zongwei Wu, Radu
  Timofte, et~al.
\newblock {NTIRE} 2025 ambient lighting normalization challenge.
\newblock In {\em Proceedings of the IEEE/CVF Conference on Computer Vision and
  Pattern Recognition (CVPR) Workshops}, 2025.

\bibitem{wan2024psc}
Fei Wan, Bingxin Xu, Weiguo Pan, and Hongzhe Liu.
\newblock Psc diffusion: patch-based simplified conditional diffusion model for
  low-light image enhancement.
\newblock {\em Multimedia Systems}, 30(4):187, 2024.

\bibitem{ntire2025lightfield}
Yingqian Wang, Zhengyu Liang, Fengyuan Zhang, Lvli Tian, Longguang Wang,
  Juncheng Li, Jungang Yang, Radu Timofte, Yulan Guo, et~al.
\newblock {NTIRE} 2025 challenge on light field image super-resolution: Methods
  and results.
\newblock In {\em Proceedings of the IEEE/CVF Conference on Computer Vision and
  Pattern Recognition (CVPR) Workshops}, 2025.

\bibitem{ssim}
Zhou Wang, A.~C. Bovik, H.~R. Sheikh, and E.~P. Simoncelli.
\newblock Image quality assessment: from error visibility to structural
  similarity.
\newblock {\em IEEE Transactions on Image Processing}, 2004.

\bibitem{yan2025hvi}
Qingsen Yan, Yixu Feng, Cheng Zhang, Guansong Pang, Kangbiao Shi, Peng Wu, Wei
  Dong, Jinqiu Sun, and Yanning Zhang.
\newblock Hvi: A new color space for low-light image enhancement.
\newblock {\em arXiv preprint arXiv:2502.20272}, 2025.

\bibitem{ntire2025reflection}
Kangning Yang, Jie Cai, Ling Ouyang, Florin-Alexandru Vasluianu, Radu Timofte,
  Jiaming Ding, Huiming Sun, Lan Fu, Jinlong Li, Chiu~Man Ho, Zibo Meng, et~al.
\newblock {NTIRE} 2025 challenge on single image reflection removal in the
  wild: Datasets, methods and results.
\newblock In {\em Proceedings of the IEEE/CVF Conference on Computer Vision and
  Pattern Recognition (CVPR) Workshops}, 2025.

\bibitem{yang2021simam}
Lingxiao Yang, Ru-Yuan Zhang, Lida Li, and Xiaohua Xie.
\newblock Simam: A simple, parameter-free attention module for convolutional
  neural networks.
\newblock In {\em International conference on machine learning}, pages
  11863--11874. PMLR, 2021.

\bibitem{lol-v2}
Wenhan Yang, Wenjing Wang, Haofeng Huang, Shiqi Wang, and Jiaying Liu.
\newblock Sparse gradient regularized deep retinex network for robust low-light
  image enhancement.
\newblock {\em IEEE Transactions on Image Processing}, 30:2072--2086, 2021.

\bibitem{yu2022towards}
Xin Yu, Peng Dai, Wenbo Li, Lan Ma, Jiajun Shen, Jia Li, and Xiaojuan Qi.
\newblock Towards efficient and scale-robust ultra-high-definition image
  demoir{\'e}ing.
\newblock In {\em European Conference on Computer Vision}, pages 646--662.
  Springer, 2022.

\bibitem{ntire2025hrdepth}
Pierluigi Zama~Ramirez, Fabio Tosi, Luigi Di~Stefano, Radu Timofte, Alex
  Costanzino, Matteo Poggi, Samuele Salti, Stefano Mattoccia, et~al.
\newblock {NTIRE} 2025 challenge on hr depth from images of specular and
  transparent surfaces.
\newblock In {\em Proceedings of the IEEE/CVF Conference on Computer Vision and
  Pattern Recognition (CVPR) Workshops}, 2025.

\bibitem{zamir2022restormer}
Syed~Waqas Zamir, Aditya Arora, Salman Khan, Munawar Hayat, Fahad~Shahbaz Khan,
  and Ming-Hsuan Yang.
\newblock Restormer: Efficient transformer for high-resolution image
  restoration.
\newblock In {\em Proceedings of the IEEE/CVF Conference on Computer Vision and
  Pattern Recognition}, pages 5728--5739, 2022.

\bibitem{mirv2}
Syed~Waqas Zamir, Aditya Arora, Salman Khan, Munawar Hayat, Fahad~Shahbaz Khan,
  Ming-Hsuan Yang, and Ling Shao.
\newblock Learning enriched features for fast image restoration and
  enhancement.
\newblock {\em IEEE transactions on pattern analysis and machine intelligence},
  45(2):1934--1948, 2022.

\bibitem{kind}
Yonghua Zhang, Jiawan Zhang, and Xiaojie Guo.
\newblock Kindling the darkness: A practical low-light image enhancer.
\newblock In {\em Proceedings of the 27th ACM international conference on
  multimedia}, pages 1632--1640, 2019.

\bibitem{zhang2023frc}
Zhao Zhang, Huan Zheng, Richang Hong, Jicong Fan, Yi Yang, and Shuicheng Yan.
\newblock Frc-net: A simple yet effective architecture for low-light image
  enhancement.
\newblock {\em IEEE Transactions on Consumer Electronics}, 70(1):3332--3340,
  2023.

\bibitem{zhou2024glare}
Han Zhou, Wei Dong, Xiaohong Liu, Shuaicheng Liu, Xiongkuo Min, Guangtao Zhai,
  and Jun Chen.
\newblock Glare: Low light image enhancement via generative latent feature
  based codebook retrieval.
\newblock In {\em European Conference on Computer Vision}, pages 36--54.
  Springer, 2024.

\bibitem{zhou2023breaking}
Han Zhou, Wei Dong, Yangyi Liu, and Jun Chen.
\newblock Breaking through the haze: An advanced non-homogeneous dehazing
  method based on fast fourier convolution and convnext.
\newblock In {\em Proceedings of the IEEE/CVF Conference on Computer Vision and
  Pattern Recognition}, pages 1894--1903, 2023.

\bibitem{Zhang2022}
W. Zou, T. Ye, W. Zheng, Y. Zhang, L. Chen, and Y. Wu.
\newblock Self-calibrated efficient transformer for lightweight
  super-resolution.
\newblock In {\em Proceedings of the IEEE/CVF Conference on Computer Vision and
  Pattern Recognition (CVPR)}, pages 930--939, 2022.

\end{thebibliography}
